\documentclass[accepted]{uai2022} 

\usepackage[american]{babel}

\usepackage{natbib} 
\usepackage{mathtools} 
\usepackage{booktabs} 
\usepackage{tikz} 

\usepackage{amssymb}
\usepackage{amsmath}
\usepackage{amsthm}
\usepackage{bbm}

\usepackage[utf8]{inputenc} 
\usepackage[T1]{fontenc}    
\usepackage{hyperref}       
\usepackage{url}            
\usepackage{amsfonts}       
\usepackage{nicefrac}       
\usepackage{microtype}      
\usepackage{xcolor}         
\usepackage{comment}  
\usepackage{graphicx}
\usepackage{subcaption}

\usepackage[ruled,vlined]{algorithm2e}
            
\usepackage{cleveref}
\crefformat{section}{\S#2#1#3} 
\crefformat{subsection}{\S#2#1#3}
\crefformat{subsubsection}{\S#2#1#3}

\newtheorem*{theorem*}{Theorem}
\newtheorem{theorem}{Theorem}[section]

\newtheorem{lemma}[theorem]{Lemma}

\DeclareMathOperator*{\argmin}{arg\,min}

\newcommand{\mP}{\mathbf{P}}

\newcommand{\x}{\mathbf{x}}
\newcommand{\y}{y}
\newcommand{\X}{\mathbf{X}}
\newcommand{\Y}{\mathbf{Y}}
\newcommand{\z}{\mathbf{z}}
\newcommand{\e}{\mathbf{e}}

\newcommand{\h}{f_h}
\newcommand{\s}{f_m}
\newcommand{\dom}{\textnormal{dom}}

\newcommand{\wt}{\widehat{\theta}}

\newcommand{\eptt}{\epsilon_{\mP_t}(\theta)}
\newcommand{\wepst}{\widehat{\epsilon}_{\mP_s}(\theta)}

\usepackage{listings}
\usepackage{fancyvrb}

\title{Toward Learning Human-aligned Cross-domain Robust Models by Countering Misaligned Features}

\author[1]{Haohan Wang}
\author[2]{Zeyi Huang}
\author[1]{Hanlin Zhang}
\author[2]{Yong Jae Lee}
\author[1,3,4]{Eric P. Xing}
\affil[1]{%
    School of Computer Science\\
    Carnegie Mellon University\\
    Pittsburgh, PA, USA
}
\affil[2]{%
    Department of Computer Sciences\\
    University of Wisconsin-Madison\\
    Madison, WI, USA
}
\affil[3]{%
    Mohamed bin Zayed University of Artificial Intelligence\\
    Abu Dhabi, United Arab Emirates
  }
\affil[4]{%
    Petuum, Inc.\\
    Pittsburgh, PA, USA
  }
  
\begin{document}
\maketitle






\begin{abstract}
  Machine learning has demonstrated 
remarkable prediction accuracy over \textit{i.i.d} data, 
but the accuracy 
often drops when tested 
with data from another distribution. 
In this paper, 
we aim to offer another view of this problem
in a perspective 
assuming 
the reason behind this accuracy drop
is the reliance of models 
on the features that are not aligned well with how a data annotator considers similar across these two datasets. 
We refer to these features as misaligned features. 
We extend the conventional generalization error bound 
to a new one 
for this setup
with the knowledge of how the misaligned
features are associated with the label.
Our analysis offers a set
of techniques for this problem, 
and these techniques are naturally linked to many previous methods in robust machine learning literature. 
We also compared the empirical strength of these methods demonstrated the performance when these previous techniques are combined, with implementation available \href{https://github.com/OoDBag/WR}{here}. 
\end{abstract}

\section{Introduction}
\begin{figure*}[t]
    \centering 
    \includegraphics[width=0.9\textwidth]{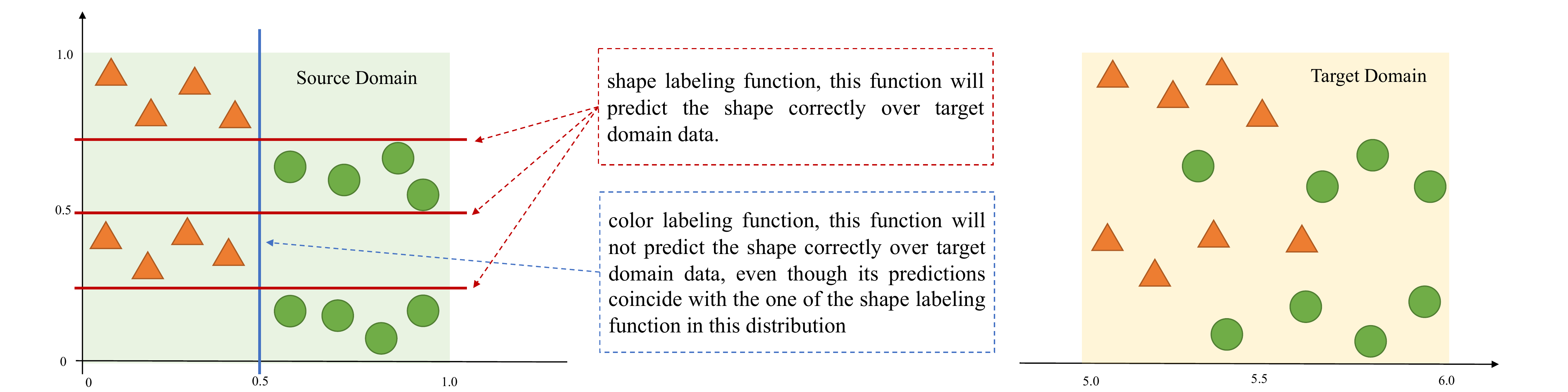}
    \caption{An illustration of the main problem focused in this paper. 
    }
    \label{fig:intro}
\end{figure*}

Machine learning, 
especially deep neural networks, 
has demonstrated remarkable empirical successes
over various applications. 
The models even
occasionally achieved
results beyond human-level performances 
over benchmark datasets
\citep[\textit{e.g.,}][]{he2015delving}. 
However, whether it is desired 
for a model to outsmart human on benchmarks remains 
an open discussion in recent years: 
indeed,
a model can create more application opportunities when it surpasses
human-level performances,
but the community also notices that the performance gain
is sometimes due to model's exploitation of the features 
meaningless to a human, 
which may lead to unexpected performance drops when the models
are tested with other datasets in practice 
that a human considers similar to the benchmark \citep{christian2020alignment}. 

One of the most famous examples of the model's exploitation of 
non-human-aligned features is probably the 
usage of snow background in ``husky vs. wolf'' image classification \citep{ribeiro2016should}. 
Briefly, 
when the model is trained to classify ``husky vs wolf,''
it notices that wolf images usually 
have a snow background and learns to use the background features. 
This example is only one of many similar discussions 
concerning that the models are using features considered futile by humans \citep[\textit{e.g.,}][]{WangGLX19,SunGTHEZMBCW19}, 
and, sometimes, the features used 
are not even perceptible to a human 
\citep{geirhos2018imagenettrained,ilyas2019adversarial,wang2020high,hermann2020origins}. 
The usage of these features might lead to a misalignment between the human and the models' understanding of the data, 
leading to a potential performance drop when the models are applied to other data that a human considers similar. 

We illustrate this challenge with a toy example in Figure~\ref{fig:intro},
where the model is trained on the source domain data 
to classify triangle vs. circle
and tested on the target domain data 
with a different marginal distribution. 
However, the color coincides 
with the shape on the source domain. 
As a result, the model might learn either the shape function
or the color one. 
The color function 
will not classify the target domain data correctly 
while the shape function can, 
but the empirical risk minimizer (ERM) cannot differentiate them 
and might learn either one, 
leading to potentially degraded performances during the test. 
As one might expect, 
whether shape or color is considered human-aligned 
is subjective depending on the task or the data 
and, in general, irrelevant to the statistical nature of the problem. 
Therefore, our remaining analysis will depend on such knowledge. 

In this paper, we aim to formalize the above challenge 
to study the learning of human-aligned models. 
In particular, we derive a new generalization error bound 
when a model is trained on one distribution
but tested on another one that human consider similar. 
As discussed previously, 
one potential challenge for this scenario
is that the model may learn 
to use some features, 
which we refer to as \emph{misaligned features},
that a human considers irrelevant. 
Corresponding to this challenge, 
our analysis will be built upon the knowledge of 
how misaligned features are associated with the label.

\section{Related Work}
\label{sec:related}
There is a recent proliferation of methods 
aiming to learn robust models 
by enforcing the models to disregard certain features. 
We consider these works direct precedents of our discussion
because these features are usually 
defined when comparing the model's performances to a human's. 
For example, the texture or background of images is probably 
the most discussed misaligned features for image classification. 
We briefly discuss these works in two main strategies. 

\paragraph{Data Augmentation}
With the knowledge of the misaligned features, 
the most effective solution is probably to augment the data
by perturbing these misaligned features.
Some recent examples of the perturbations used to train robust models 
include style transfer of images \citep{geirhos2018imagenettrained}, 
naturalistic augmentation (color distortion, noise, and blur) of images \citep{hermann2020origins}, 
other naturalistic augmentations (texture, rotation, contrast) of images \citep{wang2020squared},
interpolation of images \citep{hendrycks2019augmix}, 
syntactic transformations of sentences \citep{MahabadiBH20}, 
and across data domain \citep{ShankarPCCJS18,huang2020self,lee2021removing,huang2022two}. 

Further, as recent studies suggest that 
one reason for the adversarial vulnerability \citep{szegedy2013intriguing,goodfellow2015explaining} 
is the existence of imperceptible features correlated with the label \citep{ilyas2019adversarial,wang2020high}, 
improving adversarial robustness may also be about 
countering the model's tendency toward learning these features. 
Currently, one of the most widely accepted methods to improve adversarial robustness 
is to augment the data along the training process to maximize the training loss 
by perturbing these features within predefined robustness constraints 
(\textit{e.g.}, within $\ell_p$ norm ball) \citep{MadryMSTV18}.
While this augmentation strategy is widely referred to as adversarial training, 
for the convenience of our discussion, 
we refer to it as the
worst-case data augmentation, following the naming conventions of \citep{FawziSTF16}. 

\paragraph{Regularizing Hypothesis Space}
Another thread is to introduce inductive bias 
(\textit{i.e.}, to regularize the hypothesis space)
to force the model to discard misaligned features. 
To achieve this goal, 
one usually needs to first construct a side component 
to inform the main model about the misaligned features, 
and then to regularize the main model according to the side component. 
The construction of this side component 
usually relies on prior knowledge of what the misaligned features are.
Then, methods can be built accordingly to counter the features such as the texture of images \citep{WangHLX19,bahng2019learning}, 
the local patch of images \citep{WangGLX19}, 
label-associated keywords \citep{he2019unlearn},
label-associated text fragments \citep{MahabadiBH20}, 
and general easy-to-learn patterns of data \citep{nam2020learning}. 


In a broader scope, following the argument
that one of the main challenges of domain adaptation is to 
counter the model's tendency in learning domain-specific features \citep[\textit{e.g.},][]{GaninUAGLLML16, li2018domain}, 
some methods contributing to domain adaption
may have also progressed along the line of our interest. 
The most famous example is probably 
the domain adversarial neural network (DANN) \citep{GaninUAGLLML16}. 
Inspired by the theory of domain adaptation \citep{ben2010theory}, 
DANN trains the cross-domain generalizable neural network with the help of a side component specializing in classifying samples' domains. 
The subtle difference between this work and the ones mentioned previously is that 
this side component is not constructed with a special inductive bias 
but built as a simple network learning to classify domains with auxiliary annotations (domain IDs). 
DANN also inspires a family of methods 
forcing the model to learn auxiliary-annotation-invariant representations 
with a side component such as \citep{ghifary2016deep,rozantsev2018beyond,motiian2017unified,li2018domain,carlucci2018agnostic}. 

\paragraph{Relation to Previous Works}
The above methods solve the same human-aligned learning problems 
with two different perspectives, but we notice the same 
central theme of forcing the models to \emph{not} learn something 
according to the prior knowledge of the data or the task. 
Although this central theme has been noticed by prior works such as \citep{WangHLX19,bahng2019learning,MahabadiBH20}, 
we notice a lack of formal analysis from a task-agnostic viewpoint. 
Therefore, we continue to investigate whether we can 
contribute a principled understanding of this central theme, 
which serves as a connection of these methods and, 
potentially, a guideline for developing future methods. 
Also, we notice that many works along the domain adaptation development 
have rigorous statistical analysis \citep{ben2007analysis,ben2010theory,MansourMR09,GermainHLM16,ZhangLLJ19,dhouib2020margin}, 
and these analyses mostly focus on the alignment of the distributions. 
Our study will complement these works by investigating through the perspective of misaligned features. 
The advantages and limitations of our perspective will also be discussed. 

\section{Generalization Understanding of Human-aligned Robust Models}
\label{sec:cua}
\paragraph{Roadmap} We study the generalization error bound of human-aligned robust model in this section. We will first set up the problem of studying the generalization of the model across two distributions, whose difference mainly lies in the fact that one distribution has another labelling function (namely, the misaligned labelling function) in addition to the one that is shared across both of these distributions (\textbf{A2}). 
Then, to help quantify the error bound, we need to define the active set (features used by the function) ($\mathcal{A}(f,\x)$ in \eqref{eq:a:def}), 
the difference between the two functions ($d(\theta, f, \x)$ in \eqref{eq:d:def}), 
and an additional term to quantify whether the model learns the function if the model can map the sample correctly ($r(\theta, \mathcal{A}(f,\x))$ in \eqref{eq:r}). 
With these terms defined, we will show a formal result on the generalization error bound, which depends on how many training samples are predicted correctly when the model learns the mis-aligned samples in addition to the standard terms.

\subsection{Notations \& Background}
We consider a binary classification problem 
from feature space $\mathcal{X} \in \mathbb{R}^p$ to 
label space $\mathcal{Y} \in \{0, 1\}$. 
The distribution 
over $\mathcal{X}$ is denoted as $\mP$. 
A \emph{labeling function} $f:\mathcal{X} \rightarrow \mathcal{Y}$
is a function that
maps the feature $\x$ to its label $\y$. 
A \emph{hypothesis} or \emph{model} 
$\theta:\mathcal{X} \rightarrow \mathcal{Y}$ is also 
a function that maps the feature to the label. 
The difference in naming is only because 
we want to differentiate 
whether the function 
is a 
natural property of the space or distribution (thus called a labeling function)
or a function to estimate (thus called a hypothesis or model). 
The hypothesis space is denoted as $\Theta$. 
We use $\dom$ to denote the domain (input space) of a function, 
thus $\dom(\theta) = \mathcal{X}$.

This work studies the generalization error 
across two distributions, 
namely source and target distribution, 
denoted as $\mP_s$ and $\mP_t$, respectively. 
We are only interested when these two distributions
are, considered by a human, similar but different: 
being similar means 
there exists a \emph{human-aligned labeling function}, $\h$, 
that maps any $\x \in \mathcal{X}$ to its label
(thus the label $\y := \h(\x)$); 
being different means 
there exists a \emph{misaligned labeling function}, $\s$, 
that for any $\x \sim \mP_s$, $\s(\x) = \h(\x)$. 
This ``similar but different'' property will 
be reiterated as an assumption (\textbf{A2}) later. 
We use $(\x,\y)$ to denote a sample, and use $(\X, \Y)_\mP$ to denote a finite dataset if the features are from $\mP$ (see detailed process from \textbf{A2}).
We use $\epsilon_\mP(\theta)$ to denote the expected risk of $\theta$ 
over distribution $\mP$,
and use $\widehat{\cdot}$ to denote the estimation of the term $\cdot$
(\textit{e.g.}, the empirical risk is $\widehat{\epsilon}_\mP(\wt)$).
We use $l(\cdot,\cdot)$ to denote a generic loss function. 

For a dataset $(\X, \Y)_\mP$, 
if we train a model with
\begin{align}
    \wt = \argmin_{\theta \in \Theta}\sum_{(\x, \y)\in (\X,\Y)_\mP}l(\theta(\x), \y),
    \label{eq:train}
\end{align}
previous generalization study suggests that we can expect the error rate to be bounded as 
\begin{align}
    \epsilon_\mP(\wt) \leq \widehat{\epsilon}_\mP(\wt) + \phi(|\Theta|, n,  \delta),
    \label{eq:bound:vanilla}
\end{align}
where $\epsilon_\mP(\wt)$ and $\widehat{\epsilon}_\mP(\wt)$ respectively are 
\begin{align*}
    \epsilon_\mP(\wt) = \mathbb{E}_{\x \sim \mP}|\wt(\x)-\y|=\mathbb{E}_{\x \sim \mP}|\wt(\x)-\h(\x)|
\end{align*}
and
\begin{align*}
    \widehat{\epsilon}_\mP(\wt) = \dfrac{1}{n}\sum_{(\x, \y)\in (\X, \Y)_\mP}|\wt(\x)-\y| ,
\end{align*}
and 
$\phi(|\Theta|, n,  \delta)$ is a function 
of hypothesis space $|\Theta|$, number of samples $n$, 
and the probability when the bound holds $\delta$. 
This paper expands the discussion with this generic form that 
can relate to several discussions, each with its own assumptions. 
We refer to these assumptions as \textbf{A1}. 
\begin{itemize}
    \item [\textbf{A1}:] basic assumptions needed to derived \eqref{eq:bound:vanilla}, for example,
    \begin{itemize}
    \item when \textbf{A1} is ``$\Theta$ is finite, $l(\cdot, \cdot)$ is a zero-one loss, samples are \textit{i.i.d}'',  $\phi(|\Theta|, n, \delta)=\sqrt{(\log(|\Theta|) + \log(1/\delta))/2n}$
    \item when \textbf{A1} is ``samples are \textit{i.i.d}'', $\phi(|\Theta|, n, \delta) = 2\mathcal{R}(\mathcal{L}) + \sqrt{(\log{1/\delta})/2n}$, where $\mathcal{R}(\mathcal{L})$ stands for Rademacher complexity and $\mathcal{L} = \{l_{\theta} \,|\, \theta \in \Theta \}$, where $l_{\theta}$ is the loss function corresponding to $\theta$. 
\end{itemize}
For more information, 
we refer interested readers to relevant textbooks such as \citep{bousquet2003introduction} for formal and intuitive discussions.
\end{itemize}

\subsection{Generalization Error Bound of Human-aligned Robust Models}
Formally, we state the challenge of our human-aligned robust learning problem as the assumption:
\begin{itemize}
    \item [\textbf{A2}:] \textbf{Existence of Misaligned Features:}
    For any $\x \in \mathcal{X}$, $\y := \h(\x)$. 
    We also have a $\s$ 
    that is different from $\h$, and for $\x \sim \mP_s$, 
    $\h(\x) = \s(\x)$. 
\end{itemize}
Thus, 
the existence of $\s$ is a key challenge for
the small empirical risk over $\mP_s$ 
to be generalized to $\mP_t$, 
because 
$\theta$ that learns either $\h$ or $\s$
will lead to small source error, 
but only $\theta$ that learns $\h$ will 
lead to small target error. 
Note that $\s$ 
may not exist for an arbitrary $\mP_s$. 
In other words, 
\textbf{A2} can be interpreted to ensure the a property  
of $\mP_s$ so that $\s$, while being different from $\h$, exists for any $\x \sim \mP_s$. 

In this problem, 
$\s$ and $\h$ are not the same 
despite $\s(\x)=\h(\x)$ for any $\x \sim \mP_s$, 
and we focus on the case where the differences 
lie in the features they use. 
To describe this difference, 
we introduce the notation $\mathcal{A}(\cdot,\cdot)$, 
which denotes a set parametrized by the labeling function and the sample, 
to describe the \emph{active set} of features used by the labeling function. 
By \emph{active set}, we refer to the minimum set of features that 
a labeling function requires to map a sample to its label. 
Formally, we define 
\begin{align}
\begin{split}
    & \mathcal{A}(f,\x) = \{i | \widehat{\z}_i = \x_i\}, \quad \textnormal{where,} \\
    & \widehat{\z} = \argmin_{\z \in \dom(f), f(\z)=f(\x)} \vert\{i \vert \z_i = \x_i\}\vert, 
    \label{eq:a:def}
\end{split}
\end{align}
and $\vert\cdot\vert$ measures the cardinality. 
Intuitively, $\mathcal{A}(f,\x)$ indexes the features $f$ uses to predict $\x$. 
Although $\s(\x)=\h(\x)$,  
$\mathcal{A}(\s,\x)$ and $\mathcal{A}(\h,\x)$ can be different.
$\mathcal{A}(\s,\x)$ is the \emph{misaligned features} 
following our definition. 

Further, we define a function difference given a sample as
\begin{align}
    d(\theta, f, \x) = \max_{\z \in \dom(f): \z_{\mathcal{A}(f,\x)}=\x_{\mathcal{A}(f,\x)}} |\theta(\z) - f(\z)|,
    \label{eq:d:def}
\end{align}
where $\x_{\mathcal{A}(f,\x)}$ denotes the features of $\x$ indexed by $\mathcal{A}(f,\x)$. 
In other words, the distance describes:
given a sample $\x$, 
the maximum disagreement 
of the two functions $\theta$ and $f$ 
for all the other data $\z \in \mathcal{X}$ with a constraint that 
the features indexed by $\mathcal{A}(f,\x)$ 
are the same as those of $\x$. 
Notice that this difference is not symmetric, as the 
active set is determined by the second function. 
By definition, we have
$d(\theta, f, \x) \geq \vert \theta(\x) -f(\x)\vert$.

Also, please notice that when we use expressions such as $\z_{\mathcal{A}(f,\x)}=\x_{\mathcal{A}(f,\x)}$, 
we imply that $\mathcal{A}(f,\x)$ is the same in both LHS and RHS.
Under this premise of the notation, 
whether \eqref{eq:a:def} has a unique solution or not will not affect our main conclusion. 

In addition, one may notice the connection between $\mathcal{A}(f,\x)$ and the minimum sufficient explanation discussed previously \citep[\textit{e.g.,}][]{camburu2020struggles,yoon2018invase,carter2019made,ribeiro2018anchors}. 
While $\mathcal{A}(f,\x)$ is conceptually the same as the minimum set of features for a model to predict, we define it mathematically different.  

To continue, we introduce the following assumption:
\begin{itemize}
    \item [\textbf{A3}:] \textbf{Realized Hypothesis:} 
    Given a large enough hypothesis space $\Theta$, for any sample $(\x, \y)$, 
    for any $\theta \in \Theta$, 
    which is not a constant mapping, 
    if $\theta(\x)=\y$, then 
    $d(\theta, \h, \x)d(\theta, \s, \x)=0$
\end{itemize}

Intuitively, 
\textbf{A3} assumes $\theta$ at least
learns one labeling function
for the sample $\x$
if $\theta$ can map the $\x$ correctly. 

Finally, to describe how $\theta$ depends on the active set of $f$, we introduce the term 
\begin{align}
    r(\theta, \mathcal{A}(f,\x)) = \max_{\z_{\mathcal{A}(f,\x)} \in \dom(f)_{\mathcal{A}(f,\x)}} |\theta(\z) - \y|,
    \label{eq:r}
\end{align}
where $\z_{\mathcal{A}(f,\x)} \in \dom(f)_{\mathcal{A}(f,\x)}$ 
denotes that the features of $\z$ indexed by $\mathcal{A}(f,\x)$ are searched in the input space $\dom(f)$. 
Notice that $r(\theta, \mathcal{A}(f,\x))=1$ alone does not mean $\theta$ depends on the active set of $f$; 
it only means so when we also have $\theta(\x)=\y$ (see the formal discussion in Lemma~\ref{lemma:iff}).
In other words, $r(\theta, \mathcal{A}(f,\x))=1$ alone may not have an intuitive meaning, 
but given $\theta(\x)=\y$, $r(\theta, \mathcal{A}(f,\x))=1$ intuitively means $\theta$ learns $f$. 

With all above, we can extend the conventional generalization error bound with a new term as follows:
\begin{theorem}[The Curse of Universal Approximation]
With Assumptions \textbf{A1}-\textbf{A3}, $l(\cdot, \cdot)$ is a zero-one loss, with probability as least $1 - \delta$, we have 
\begin{align}
    \eptt \leq \wepst + c(\theta) + \phi(|\Theta|, n,  \delta)
\end{align}
where 
\begin{align*}
c(\theta) =  \dfrac{1}{n}\sum_{(\x, \y) \in (\X, \Y)_{\mP_s}} \mathbb{I}[\theta(\x)=\y]r(\theta, \mathcal{A}(\s,\x)).
\end{align*}
\label{thm:cua}
\end{theorem}

$\mathbb{I}[\cdot]$ is a function that returns $1$ if the 
condition $\cdot$ holds and 0 otherwise. 
As $\theta$ may learn $\s$, 
$\wepst$ is not representative of $\eptt$; 
thus, we introduce $c(\theta)$ to account for the discrepancy. 
Intuitively, $c(\theta)$ quantifies 
the samples that are correctly predicted, 
but only because the $\theta$ learns $\s$ for that sample. 
$c(\theta)$
depends on the knowledge of 
$\s$. 

We name Theorem~\ref{thm:cua}
\emph{the curse of universal approximation}
to highlight the fact 
that the existence of $\s$ is not always obvious, 
but the models can usually learn it nonetheless \citep{wang2020high} . 
Even in a well-curated dataset
that does not seemingly have misaligned features, 
modern models 
might still use some features not understood by human.
This argument may also align with
recent discussions suggesting 
that
reducing the model complexity 
can improve cross-domain generalization \citep{chuang2020estimating}. 

\subsection{In Comparison to the View of Domain Adaptation}
\label{sec:cua:da}

We continue to compare Theorem~\ref{thm:cua} with 
understandings of domain adaptation. 
Conveniently, several domain adaptation analyses \citep{ben2007analysis,ben2010theory,MansourMR09,GermainHLM16,ZhangLLJ19,dhouib2020margin} can be sketched in the following form:
\begin{align}
    \eptt \leq \wepst + D_\Theta(\mP_s, \mP_t) + \lambda + \phi'(|\Theta|, n, \delta)
\end{align}
where $D_\Theta(\mP_s, \mP_t)$ quantifies the differences between the two distributions; 
$\lambda$ describes the nature of the problem 
and usually involves non-estimable terms about the problem. 

For example, \cite{ben2010theory} formalized the difference as $\Theta$-divergence, and described the corresponding empirical term as (with $\Theta\Delta\Theta$ denoting the set of disagreement between two hypotheses in $\Theta$): 
\begin{align}
\begin{split}
    D_\Theta(\mP_s, \mP_t) = & 
    1 - \min_{\theta \in \Theta\Delta\Theta}(\dfrac{1}{n}\sum_{\x:\theta(\x)=0}\mathbb{I}[\x \in (\X, \Y)_{\mP_s}] \\
    &+ \dfrac{1}{n}\sum_{\x:\theta(\x)=1}\mathbb{I}[\x \in (\X, \Y)_{\mP_t}]).
    \label{eq:h-divergence}
\end{split}
\end{align}
Also, \cite{ben2010theory} formalized
$\lambda = \epsilon_{\mP_t}(\theta^\star) + \epsilon_{\mP_s}(\theta^\star)$, 
where 
$\theta^\star = \argmin_{\theta\in\Theta}\epsilon_{\mP_t}(\theta) + \epsilon_{\mP_s}(\theta)$, 

In our discussion, as we assume the $\h$ applies to any $\x \in \mathcal{X}$ (according to \textbf{A2}), $\lambda=0$ as long as the hypothesis space is large enough. Therefore, the comparison mainly lies in comparing $c(\theta)$ and $D_\Theta(\mP_s, \mP_t)$.

To compare them, 
we need an extra assumption:
\begin{itemize}
    \item [\textbf{A4}:] \textbf{Sufficiency of Training Samples}
    for the two finite datasets in the study, 
    \textit{i.e.}, $(\X,\Y)_{\mP_s}$ and $(\X,\Y)_{\mP_t}$, 
    for any $\x \in (\X,\Y)_{\mP_t}$, 
    there exists one or many $\z \in (\X,\Y)_{\mP_s}$ such that 
    \begin{align}
        \x \in \{\x'| \x' \in \mathcal{X} \; \textnormal{and} \; 
        \x'_{\mathcal{A}(f_h,\z)} = \z_{\mathcal{A}(f_h,\z)}
        \}
    \end{align}
\end{itemize}

\textbf{A4} intuitively means
the finite training dataset needs to be diverse enough to 
describe the concept that needs to be learned. 
For example, imagine building a classifier to classify mammals \textit{vs.} fishes from the distribution of photos to that of sketches, 
we cannot expect the classifier to do anything good on dolphins if dolphins only appear in the test sketch dataset. 
\textbf{A4} intuitively regulates that
if dolphins will appear in the test sketch dataset, 
they must also appear in the training dataset. 

Now, 
in comparison to \citep{ben2010theory},
we have
\begin{theorem}
With Assumptions \textbf{A2}-\textbf{A4}, 
and if $1 - f_h \in \Theta$, 
we have
\begin{align}
\begin{split}
  c(\theta) \leq & D_\Theta(\mP_s, \mP_t) \\& + \dfrac{1}{n}\sum_{(\x,\y) \in (\X, \Y)_{\mP_t}} \mathbb{I}[\theta(\x)=\y]r(\theta, \mathcal{A}(f_m,\x))
\end{split}
\end{align}
where 
\begin{align*}
    c(\theta) =  \dfrac{1}{n}\sum_{(\x,\y) \in (\X, \Y)_{\mP_s}} \mathbb{I}[\theta(\x)=\y]r(\theta, \mathcal{A}(f_m,\x))
\end{align*}
and $D_\Theta(\mP_s, \mP_t)$ is defined as in~\eqref{eq:h-divergence}. 
\label{thm:comparison}
\end{theorem}

$q(\theta) := \frac{1}{n}\sum_{(\x,\y) \in (\X, \Y)_{\mP_t}} \mathbb{I}[\theta(\x)=\y] r(\theta, \mathcal{A}(f_m,\x))$, which intuitively means that 
if $\theta$ learns $f_m$, 
how many samples $\theta$ can coincidentally predict correctly  over the finite target set
used to estimate $D_\Theta(\mP_s, \mP_t)$. 
For sanity check, 
if we replace $(\X, \Y)_{\mP_t}$ with $(\X, \Y)_{\mP_s}$, 
$D_\Theta(\mP_s, \mP_t)$ will be evaluated at 0 as it cannot differentiate two identical datasets, 
and $q(\theta)$ will be the same as $c(\theta)$. 
On the other hand, 
if no samples from $(\X, \Y)_{\mP_t}$
can be mapped correctly with $f_m$ (coincidentally), 
$q(\theta)=0$ and 
$c(\theta)$ will be a lower bound of $D_\Theta(\mP_s, \mP_t)$. 

The value of Theorem~\ref{thm:comparison} 
lies in the fact that 
for an arbitrary target dataset $(\X,\Y)_{\mP_t}$, 
no samples out of which can be predicted correctly 
by learning $f_m$ (a situation likely to occur for arbitrary datasets since $f_m$ is unlikely to be shared across the source dataset and any arbitrary target dataset), 
$c(\theta)$ will always be a lower bound of $D_\Theta(\mP_s, \mP_t)$. 

Further, 
when Assumption \textbf{A4} does not hold, 
we are unable to derive a clear relationship between 
$c(\theta)$ and $D_\Theta(\mP_s, \mP_t)$. 
The difference is mainly raised as a matter of fact that, 
intuitively, 
we are only interested in the problems that are ``solvable'' 
(\textbf{A4}, \textit{i.e.}, hypothesis that used to reduce the test error in target distribution can be learned from the finite training samples) 
but ``hard to solve'' 
(\textbf{A2}, \textit{i.e.}, another labeling function, namely $f_m$, 
exists for features sampled from the source distribution only), 
while $D_\Theta(\mP_s, \mP_t)$ estimates the divergence of two arbitrary distributions.

\subsection{Estimation of the Discrepancy}
The estimation of $c(\theta)$ mainly 
involves two challenges: 
the requirement of the knowledge of $\s$ 
and the computational cost to search over the entire space $\mathcal{X}$. 

The first challenge is unavoidable by definition
because the human-aligned learning has to be built upon 
the prior knowledge of what labeling function a human considers similar (what $\h$ is)
or its opposite (what $\s$ is). 
Fortunately, 
as discussed in Section~\ref{sec:related}, 
the methods are usually developed with prior knowledge of what the misaligned features are, 
suggesting that we may often directly have the knowledge.

The second challenge is about the computational cost to search, 
and the community has several techniques to help reduce the burden. 
For example, the search can be terminated 
once $r(\theta, \mathcal{A}(\s,\x))$ is evaluated as $1$ 
(\textit{i.e.}, once we find a perturbation of misaligned features that alters the prediction).
This procedure is similar to how adversarial attack \citep{goodfellow2015explaining} 
is used to evaluate the robustness of models. 
To further reduce the computational cost, 
one can also generate out-of-domain data by perturbing misaligned features beforehand
and use these fixed data to test models. 
Using fixed data to evaluate might not be as accurate as 
using a search process, 
but sometimes, it can be good enough to reveal some interesting properties of the models \citep{Jo2017, geirhos2018imagenettrained, wang2020high}.

\section{Methods to Learn Human-aligned Robust Models}
\label{sec:robust}

We continue to study how our analytical results above can lead to 
practical methods to learn human-aligned robust models. 
We first show that our discussion
can naturally connect to existing methods for robust machine learning 
discussed in Section~\ref{sec:related}. 

Theorem~\ref{thm:cua} suggests that
training a human-aligned robust model amounts to training 
for small $c(\theta)$ and small empirical error (\textit{i.e.}, $\wepst$). 

\subsection{Worst-case Training}
To simplify the notation, we define $\mathcal{Q}(\x):= \{ \x_{\mathcal{A}(\s,\x)} \in \dom(\s)_{\mathcal{A}(\s,\x)} \}$. 
We can consider the upper bound of $c(\theta)$
\begin{align}
\begin{split}
    c(\theta) 
    \leq &\dfrac{1}{n}\sum_{(\x,\y)\in (\X,\Y)} r(\theta, \mathcal{A}(\s,\x)) \\
    = & \dfrac{1}{n}\sum_{(\x,\y)\in (\X,\Y)} \max_{\z \in \mathcal{Q}(\x)} |\theta(\z) - \y|,
    \label{eq:method:worst}
\end{split}
\end{align}
which intuitively means that 
instead of $c(\theta)$ that studies only the correct predictions because $\theta$ learns $\s$, 
now we study any predictions because $\theta$ learns $\s$. 

Further, as 
\begin{align*}
    |\theta(\x) - \y| \leq \max_{\z \in \mathcal{Q}(\x)} |\theta(\z) - \y|, 
\end{align*}
a model with minimum $\eqref{eq:method:worst}$ naturally means 
the model will have a minimum empirical loss. 
Therefore, we can train for a small $\eqref{eq:method:worst}$, which likely leads to the model with a small empirical loss. 
Therefore, 
after we replace $|\theta(\x) - \y|$ with a generic loss term $\ell(\theta(\x),\y)$,
we can directly train a model with
\begin{align}
    \min_{\theta \in \Theta} \dfrac{1}{n}\sum_{(\x,\y)\in (\X,\Y)}  \max_{\z \in \mathcal{Q}(\x)} \ell(\theta(\z),\y)
    \label{eq:method:worst-da-result}
\end{align}
to get a model with small $c(\theta)$ and small empirical error. 

The above method is to augment the data by perturbing the misaligned features to maximize the training loss 
and solve the optimization problem with the augmented data. 
This method is the worst-case data augmentation method \citep{FawziSTF16} we discussed previously,
and is also closely connected to one of the most widely accepted methods 
for the adversarial robust problem, namely the adversarial training \citep{MadryMSTV18}.

While the above result shows that a method for learning human-aligned robust models is in mathematical connection to the worst-case data augmentation, 
in practice, a general application of this method will require some additional assumptions. 
The detailed discussions of these are in the appendix. 
\label{sec:worst-da}

We continue from the RHS of \eqref{eq:method:worst} to discuss another reformulation by reweighting sample losses for optimization, which leads to:
\begin{align}
    \dfrac{1}{n}\sum_{(\x,\y)\in (\X,\Y)} \max_{\z \in \mathcal{Q}(\x)} \lambda(\z)|\theta(\z) - \y|
    \label{eq:method:worst2}
\end{align}

The conditions (assumptions) that we need for $c(\theta) \leq$ the LHS of \eqref{eq:method:worst2} is discussed in the appendix. 
Now, we will continue with  
\begin{align}
    c(\theta) \leq \dfrac{1}{n}\sum_{(\x,\y)\in (\X,\Y)} \max_{\z \in \mathcal{Q}(\x)} \lambda(\z)|\theta(\z) - \y|
    \label{eq:method:worst2:2}
\end{align}

When \eqref{eq:method:worst2:2} holds, replacing $|\theta(\z) - \y|$ with a generic loss $\ell(\theta(\z),\y)$ and minimizing it is another direction of learning robust models, which corresponds to distributionally robust optimization (DRO) \citep{ben2013robust, duchi2021statistics}. 

Further, depends on implementations of $\lambda(\x)$, 
DRO has been implemented with different concrete solutions, 
sometimes with structural assumptions \citep{hu2018does}, 
such as
\begin{itemize}
    \item Adversarially reweighted learning (ARL) \citep{NEURIPS2020_07fc15c9}
    uses another model $\phi: \mathcal{X} \times \mathcal{Y} \rightarrow[0,1]$ to identify samples with misaligned features that cause high losses of model $\theta$ and defines \begin{align*}
        \lambda(\x)=1+\vert(\X, \Y)\vert \cdot \frac{\phi\left(\x\right)}{\sum_{(\x,\y) \in (\X, \Y)} \phi\left(\x\right)}
    \end{align*}
    \item Learning from failures (LFF) \citep{nam2020learning} also trains another model $\phi$ by amplifying its early-stage predictions and defines 
    \begin{align}
        \lambda(\x)=\frac{\ell\left(\phi(\x), \y\right)}{\ell\left(\phi(\x), \y\right)+\ell\left(\theta(\x), \y\right)}
    \end{align}
    \item Group DRO \citep{Sagawa*2020Distributionally} assumes the availability of the structural partition of the samples, and defines the weight of samples at partition $\mathbf{g}$ as
    \begin{align}
     \lambda(\x)= \dfrac{\exp \left(\ell\left(\theta(\x), \y)\right)\right)}{\sum_{(\z,\y) \in (\X, \Y)_\mathbf{g}} \exp \left(\ell\left(\theta(\z), \y)\right)\right)},
    \end{align}
    if $(\x, \y) \in (\X, \Y)_\mathbf{g}$, samples of partition $\mathbf{g}$
\end{itemize}
These discussions are expanded in the appendix.

\subsection{Regularizing the Hypothesis Space}
Connecting our theory to the other main thread is little bit tricky, 
as we need to extend the model to an encoder/decoder structure, where we use $e_\theta$ and $d_\theta$ to denote them respectively. 
Thus, by definition of classification models, we have $\theta(\x) = d_\theta(e_\theta(\x))$.  
Further,
we define $\s'$ as the equivalent of $\s$ with the only difference is that 
$\s'$ operates
on the representations $e_\theta(\x)$. 
With the setup, 
optimizing the empirical loss and $c(\theta)$ leads to (details in the appendix):
\begin{align}
    \min_{d_\theta, e_\theta}\dfrac{1}{n}\sum_{(\x,\y)\in (\X,\Y)}\ell(d_\theta(e_\theta(\x)), \y) - \ell(\s'(e_\theta(\x)), \y), 
    \label{eq:regularization}
\end{align}
which is highly related to methods used to learn auxiliary-annotation-invariant representations, 
and the most popular example of these methods is probably DANN \citep{GaninUAGLLML16}. 

Then, the question left is how to get $\s'$. 
We can design a specific architecture given the prior knowledge of the data, then $\s'$ can be directly estimated through 
\begin{align}
    \min_{\s'} \dfrac{1}{n}\sum_{(\x,\y)\in (\X,\Y)} \ell(\s'(e_\theta(\x)),\y), 
    \label{eq:regularization:1}
\end{align}
which connects to several methods in Section~\ref{sec:related}, such as \citep{WangGLX19, bahng2019learning}. 
Alternatively, we can estimate $\s'$ with additional annotations (\textit{e.g.}, domain ids, batch ids \textit{etc}), then we can estimate the model by (with $\mathbf{t}$ denoting the additional annotation) 
\begin{align}
    \min_{\s'} \dfrac{1}{n}\sum_{(\x,\mathbf{t})\in (\X,\mathbf{T})} \ell(\s'(e_\theta(\x)),\mathbf{t}),
    \label{eq:regularization:2}
\end{align}
which connects to methods in domain adaptation literature such as \citep{GaninUAGLLML16,li2018domain}. 


\subsection{A New Heuristic: Worst-case Training with Regularized Hypothesis Space}

Our analysis showed that optimizing for small $c(\theta)$ naturally connects to one of the two mainstream families of methods used to train robust models in the literature, 
which naturally inspires us to invent a new method by combining these two directions. 
The intuition behind this design rationale is to incorporate the empirical strength of each of these methods together by directly combining the major components of these methods.

Therefore, we introduce a new heuristic that combines
the worst-case training \eqref{eq:method:worst} 
and the regularization method \eqref{eq:regularization} and \eqref{eq:regularization:2}, for which, whether the samples are originally from $(\X, \Y)$ or generated along the training will serve as the additional annotation $\mathbf{t}$. 

\begin{algorithm}
\small 
\SetAlgoLined
\KwResult{$\theta^I$}
\textbf{Input:} total iterations $I$, $(\X,\Y)$\; 
 initialize $\theta^{(0)}$, $i=1$\;
 \While{$i \leq I$}{
 \For {sample $(\x,\y)$}{
    assign additional label $\mathbf{t}_\x=0$ for $\x$\;
    sample $\z \in Q(\x)$ that maximizes $\ell(\theta(\x), \y)$\;
    assign additional label $\mathbf{t}_\z=1$ for $\z$\;
    update $f'_m$ with \eqref{eq:regularization:2}\;
    update $\theta$ with \eqref{eq:regularization}\;
    update $\theta$ with $\z$ with the equivalence of \eqref{eq:regularization} as 
    $
        \min_{d_\theta, e_\theta}\ell(d_\theta(e_\theta(\z)), \y) - \ell(\s'(e_\theta(\z)), \y), 
    $
 }
 }
 \caption{worst-case training with regularized hypothesis space}
 \label{algorithm:main}
\end{algorithm}  

In particular, our heuristic is illustrated with Algorithm~\ref{algorithm:main}. 
In practice, we will also introduce a hyperparamter to balance the two losses in \eqref{eq:regularization}.

\section{Experiments}
\label{sec:exp}

\begin{table*}[t]
\small 
\centering 
\begin{tabular}{cccccccccccc}
\hline
 & Vanilla & SN & LM & RUBi & ReBias & Mixup & Cutout & AugMix & WT & Reg & WR \\ \hline
Standard Acc. & 90.80 & 88.40 & 67.90 & 90.50 & 91.90 & 92.50 & 91.20 & 92.90 & 92.50 & 93.10 & \textbf{93.30} \\
Weighted Acc. & 88.80 & 86.60 & 65.90 & 88.60 & 90.50 & 91.20 & 90.30 & 91.70 & 91.30 & \textbf{92.20} & 92.00 \\
ImageNet-A & 24.90 & 24.60 & 18.80 & 27.70 & 29.60 & 29.10 & 27.30 & \textbf{31.50} & 28.50 & 30.00 & 29.60 \\
ImageNet-Sketch & 41.10 & 40.50 & 36.80 & 42.30 & 41.80 & 40.60 & 38.70 & 41.40 & 43.00 & 42.50 & \textbf{43.20} \\
average & 61.40 & 60.03 & 47.35 & 62.28 & 63.45 & 63.35 & 61.88 & 64.38 & 63.83 & 64.45 & \textbf{64.53} \\ \hline
\end{tabular}
\caption{Results comparison on nine super-class ImageNet classification. }
\label{tab:miniimagenet}
\end{table*}

We presented the theory supporting experiments in Appendix, and discuss performance competing results here. 

To test the performance of our new heuristic, 
we compare our methods on a fairly recent and strong baseline. 
In particular, 
we follow the setup of a direct precedent of our work \citep{bahng2019learning}
to compare the models for a nine super-class ImageNet classification \citep{ilyas2019adversarial} with class-balanced strategies. 
Also, we follow \citep{bahng2019learning}
to report standard accuracy, 
weighted accuracy, a scenario where
samples with unusual texture are weighted more, 
and accuracy over ImageNet-A \citep{hendrycks2019natural},
a collection of 
failure cases for most ImageNet trained models. 
Additionally, we also report 
the performance over ImageNet-Sketch \citep{WangGLX19}, 
an independently collected ImageNet test set
with only sketch images. 

We test our method with the pipeline made available by \citep{bahng2019learning}, 
and we compare with the vanilla network, 
and several methods that are designed to reduce the texture bias:
including 
StylisedIN (SN) \citep{geirhos2018imagenettrained}, 
LearnedMixin (LM) \citep{clark2019don}, 
RUBi \citep{cadene2019rubi},
and ReBias \citep{bahng2019learning}, 
several other baselines proved effective in learning robust models, 
such as Mix-up \citep{zhang2017mixup},
Cutout \citep{devries2017improved}, 
AugMix \cite{hendrycks2019augmix},
In addition, we compared our worst-case training (WT), 
regularization (Reg), 
and the introduced heuristic (WR). 
For our methods, we follow the observations in \citep{wang2020high} 
suggesting the relationship between frequency-based perturbation and the model's performance,
and design the augmentation of frequency-based perturbation with different radii. 

We report the results in Table~\ref{tab:miniimagenet}. 
Our results suggest that, 
while the augmentation method we used is much simpler 
than the ones used in AugMix, 
our empirical results are fairly strong in comparison. 
With simple perturbation inspired from \citep{wang2020high}, 
our new heuristic outperforms other methods in average 
on these four test scenarios.

\section{Discussion}
\label{sec:diss}
Before we conclude, we would like to devote a section to discuss several topics more broadly related to this paper. 

\textbf{Human-aligned machine learning may not be solvable in general without prior knowledge.}
Following our notations in this paper, for any two functions $f_1$ and $f_2$, it is human, instead of any statistical properties, that decides whether $\h=f_1$ or $\h=f_2$. 
This remark is a restatement of our motivating example in Figure~\ref{fig:intro}. 
Our proposed method forgoes the requirement of prior knowledge
and is validated empirically on certain benchmark datasets. 

\textbf{Do all the model's understandings of the data have to be aligned with a human's?}
Probably no.
As we have discussed in the preceding sections, 
we agree that there are also scenarios where it is beneficial for models' perception to outperform a human's. 
For example, we may expect the models to outperform the human vision system when applied to make a scientific discovery at a molecule level. 
This paper investigates these questions for the scenarios where the alignment is essential. 

\textbf{In practice, there is probably more than one source of misaligned features.}
We aim to contribute a principled understanding of 
the problem, starting with its basic form.
The extension of our analysis to multiple sources of misaligned features
is considered a future direction. 

\textbf{Differences between overfitting and non-human-aligned} 
A critical difference is that 
overfitting can typically be observed empirically 
with a split of train and test datasets, 
while learning the misaligned features 
is usually not observed because the misaligned features
can be true across the train and test data split. 

\textbf{Other related works}
There is also a proliferation of works 
that aim to improve the robustness of machine learning methods
from a data perspective, 
such as the methods developed to counter 
spurious correlations \citep{vigen2015spurious}, 
confounding factors \citep{mcdonald2014confounding}, 
or dataset bias \citep{torralba2011unbiased}. 
We believe how our analysis is statistically connected to these topics
is also an interesting future direction. 
Further, there is also an active line of research aiming to align the human and models' perception of data 
by studying how humans process the images \citep{KubiliusSHMRIKB19,MarblestoneWK16,nayebi2017biologically,lindsay2018biological,BWCMRBSPT19,DapelloMSGCD20}. 

In addition, discussion of how human annotation will help the models 
to generalize in non-trivial test scenarios has also been explored. 
For example, \citep{ross2017right} built expert annotation into the model 
to regularize the explanation of the models to counter the model's tendency in learning misaligned features. 
The study has been extended with multiple follow-ups to introduce human-annotation into the interpretation of the models \citep{schramowski2020making,teso2019explanatory,lertvittayakumjorn2020find}, 
and shows that the human's knowledge will help model's learning the concepts that can generalize 
in non-i.i.d scenarios. 


\section{Conclusion}
\label{sec:con}
In this paper, 
we built upon the importance of learning human-aligned model
and studied the generalization properties of a model
for the goal of the alignment between the human and the model. 
We extended the widely-accepted generalization error bound 
with an additional term for the differences 
between the human and the model, 
and this new term relies on 
how the misaligned features are associated with the label. 
Optimizing for small empirical loss and small this term 
will lead to a model that is better aligned to humans. 
Thus, 
our analysis naturally offers a set of methods to this problem. 
Interestingly, 
these methods are closely connected to the established methods 
in multiple topics regarding robust machine learning. 
Finally, 
by noticing our analysis can link to two mainstream families of methods of learning robust models, 
we propose a new heuristic of combining them. 
In a fairly advanced experiment, we demonstrate the empirical strength of our new method.

\subsubsection*{Acknowledgement}
This work was supported in part by NIH R01GM114311, NIH P30DA035778, NSF IIS1617583, NSF CAREER IIS-2150012 and IIS-2204808.  

\bibliography{ref}

\begin{thebibliography}{67}
\providecommand{\natexlab}[1]{#1}
\providecommand{\url}[1]{\texttt{#1}}
\expandafter\ifx\csname urlstyle\endcsname\relax
  \providecommand{\doi}[1]{doi: #1}\else
  \providecommand{\doi}{doi: \begingroup \urlstyle{rm}\Url}\fi

\bibitem[Bahng et~al.(2019)]{bahng2019learning}
Hyojin Bahng et~al.
\newblock Learning de-biased representations with biased representations.
\newblock \emph{arXiv:1910.02806}, 2019.

\bibitem[Ben-David et~al.(2007)]{ben2007analysis}
Shai Ben-David et~al.
\newblock Analysis of representations for domain adaptation.
\newblock In \emph{NeurIPS}, pages 137--144, 2007.

\bibitem[Ben-David et~al.(2010)]{ben2010theory}
Shai Ben-David et~al.
\newblock A theory of learning from different domains.
\newblock \emph{Machine learning}, 79\penalty0 (1-2):\penalty0 151--175, 2010.

\bibitem[Ben-Tal et~al.(2013)]{ben2013robust}
Aharon Ben-Tal et~al.
\newblock Robust solutions of optimization problems affected by uncertain
  probabilities.
\newblock \emph{Management Science}, 59\penalty0 (2):\penalty0 341--357, 2013.

\bibitem[Bousquet et~al.(2003)]{bousquet2003introduction}
Olivier Bousquet et~al.
\newblock Introduction to statistical learning theory.
\newblock In \emph{Summer School on Machine Learning}, pages 169--207.
  Springer, 2003.

\bibitem[Cadene et~al.(2019)]{cadene2019rubi}
Remi Cadene et~al.
\newblock Rubi: Reducing unimodal biases for visual question answering.
\newblock In \emph{NeurIPS}, pages 841--852, 2019.

\bibitem[Camburu et~al.(2020)]{camburu2020struggles}
Oana-Maria Camburu et~al.
\newblock The struggles of feature-based explanations: Shapley values vs.
  minimal sufficient subsets.
\newblock \emph{arXiv:2009.11023}, 2020.

\bibitem[Carlucci et~al.(2018)]{carlucci2018agnostic}
Fabio~M Carlucci et~al.
\newblock Agnostic domain generalization.
\newblock \emph{arXiv:1808.01102}, 2018.

\bibitem[Carter et~al.(2019)]{carter2019made}
Brandon Carter et~al.
\newblock What made you do this? understanding black-box decisions with
  sufficient input subsets.
\newblock In \emph{The 22nd AISTATS}, pages 567--576. PMLR, 2019.

\bibitem[Christian(2020)]{christian2020alignment}
Brian Christian.
\newblock \emph{The Alignment Problem: Machine Learning and Human Values}.
\newblock WW Norton \& Company, 2020.

\bibitem[Chuang et~al.(2020)]{chuang2020estimating}
Ching-Yao Chuang et~al.
\newblock Estimating generalization under distribution shifts via
  domain-invariant representations.
\newblock \emph{arXiv:2007.03511}, 2020.

\bibitem[Clark et~al.(2019)]{clark2019don}
Christopher Clark et~al.
\newblock Don't take the easy way out: Ensemble based methods for avoiding
  known dataset biases.
\newblock \emph{arXiv:1909.03683}, 2019.

\bibitem[Dapello et~al.(2020)]{DapelloMSGCD20}
Joel Dapello et~al.
\newblock Simulating a primary visual cortex at the front of cnns improves
  robustness to image perturbations.
\newblock In \emph{NeurIPS 2020}, 2020.

\bibitem[DeVries and Taylor(2017)]{devries2017improved}
Terrance DeVries and Graham~W Taylor.
\newblock Improved regularization of convolutional neural networks with cutout.
\newblock \emph{arXiv:1708.04552}, 2017.

\bibitem[Dhouib et~al.(2020)]{dhouib2020margin}
Sofien Dhouib et~al.
\newblock Margin-aware adversarial domain adaptation with optimal transport.
\newblock In \emph{Thirty-seventh ICML}, 2020.

\bibitem[Duchi et~al.(2021)]{duchi2021statistics}
John~C Duchi et~al.
\newblock Statistics of robust optimization: A generalized empirical likelihood
  approach.
\newblock \emph{Mathematics of Operations Research}, 2021.

\bibitem[Fawzi et~al.(2016)]{FawziSTF16}
Alhussein Fawzi et~al.
\newblock Adaptive data augmentation for image classification.
\newblock In \emph{ICIP 2016}. {IEEE}, 2016.

\bibitem[Ganin et~al.(2016)]{GaninUAGLLML16}
Yaroslav Ganin et~al.
\newblock Domain-adversarial training of neural networks.
\newblock \emph{J. Mach. Learn. Res.}, 17:\penalty0 59:1--59:35, 2016.

\bibitem[Geirhos et~al.(2019)]{geirhos2018imagenettrained}
Robert Geirhos et~al.
\newblock Imagenet-trained {CNN}s are biased towards texture; increasing shape
  bias improves accuracy and robustness.
\newblock In \emph{ICLR}, 2019.

\bibitem[Germain et~al.(2016)]{GermainHLM16}
Pascal Germain et~al.
\newblock A new pac-bayesian perspective on domain adaptation.
\newblock In \emph{ICML, 2016}, volume~48, pages 859--868, 2016.

\bibitem[Ghifary et~al.(2016)]{ghifary2016deep}
Muhammad Ghifary et~al.
\newblock Deep reconstruction-classification networks for unsupervised domain
  adaptation.
\newblock In \emph{ECCV}, pages 597--613. Springer, 2016.

\bibitem[Goodfellow et~al.(2015)]{goodfellow2015explaining}
Ian~J Goodfellow et~al.
\newblock Explaining and harnessing adversarial examples (2014).
\newblock In \emph{ICLR}, 2015.

\bibitem[He et~al.(2019)]{he2019unlearn}
He~He et~al.
\newblock Unlearn dataset bias in natural language inference by fitting the
  residual.
\newblock \emph{arXiv:1908.10763}, 2019.

\bibitem[He et~al.(2015)]{he2015delving}
Kaiming He et~al.
\newblock Delving deep into rectifiers: Surpassing human-level performance on
  imagenet classification.
\newblock In \emph{ICCV}, pages 1026--1034, 2015.

\bibitem[Hendrycks et~al.(2019{\natexlab{a}})]{hendrycks2019augmix}
Dan Hendrycks et~al.
\newblock Augmix: A simple data processing method to improve robustness and
  uncertainty.
\newblock In \emph{ICLR}, 2019{\natexlab{a}}.

\bibitem[Hendrycks et~al.(2019{\natexlab{b}})]{hendrycks2019natural}
Dan Hendrycks et~al.
\newblock Natural adversarial examples.
\newblock \emph{arXiv:1907.07174}, 2019{\natexlab{b}}.

\bibitem[Hermann and Kornblith(2020)]{hermann2020origins}
Katherine Hermann and Simon Kornblith.
\newblock The origins and prevalence of texture bias in convolutional neural
  networks.
\newblock \emph{NeurIPS}, 33, 2020.

\bibitem[Hu et~al.(2018)]{hu2018does}
Weihua Hu et~al.
\newblock Does distributionally robust supervised learning give robust
  classifiers?
\newblock In \emph{ICML}, pages 2029--2037. PMLR, 2018.

\bibitem[Huang et~al.(2020)]{huang2020self}
Zeyi Huang et~al.
\newblock Self-challenging improves cross-domain generalization.
\newblock In \emph{ECCV}. Springer, 2020.

\bibitem[Huang et~al.(2022)]{huang2022two}
Zeyi Huang et~al.
\newblock The two dimensions of worst-case training and their integrated effect
  for out-of-domain generalization.
\newblock In \emph{CVPR}, 2022.

\bibitem[Ilyas et~al.(2019)]{ilyas2019adversarial}
Andrew Ilyas et~al.
\newblock Adversarial examples are not bugs, they are features.
\newblock In \emph{NeurIPS}, pages 125--136, 2019.

\bibitem[Jo and Bengio(2017)]{Jo2017}
Jason Jo and Yoshua Bengio.
\newblock Measuring the tendency of cnns to learn surface statistical
  regularities.
\newblock \emph{CoRR}, abs/1711.11561, 2017.

\bibitem[Kubilius et~al.(2019)]{KubiliusSHMRIKB19}
Jonas Kubilius et~al.
\newblock Brain-like object recognition with high-performing shallow recurrent
  anns.
\newblock In \emph{NeurIPS 2019}, pages 12785--12796, 2019.

\bibitem[Lahoti et~al.(2020)]{NEURIPS2020_07fc15c9}
Preethi Lahoti et~al.
\newblock Fairness without demographics through adversarially reweighted
  learning.
\newblock In \emph{NeurIPS}, volume~33, pages 728--740, 2020.

\bibitem[Lee et~al.(2021)]{lee2021removing}
Saehyung Lee et~al.
\newblock Removing undesirable feature contributions using out-of-distribution
  data.
\newblock \emph{arXiv:2101.06639}, 2021.

\bibitem[Lertvittayakumjorn et~al.(2020)]{lertvittayakumjorn2020find}
Piyawat Lertvittayakumjorn et~al.
\newblock Find: human-in-the-loop debugging deep text classifiers.
\newblock \emph{arXiv:2010.04987}, 2020.

\bibitem[Li et~al.(2018)]{li2018domain}
Haoliang Li et~al.
\newblock Domain generalization with adversarial feature learning.
\newblock In \emph{CVPR}, 2018.

\bibitem[Li et~al.(2019)]{BWCMRBSPT19}
Zhe Li et~al.
\newblock Learning from brains how to regularize machines.
\newblock In \emph{NeurIPS 2019}, pages 9525--9535, 2019.

\bibitem[Lindsay and Miller(2018)]{lindsay2018biological}
Grace~W Lindsay and Kenneth~D Miller.
\newblock How biological attention mechanisms improve task performance in a
  large-scale visual system model.
\newblock \emph{ELife}, 7:\penalty0 e38105, 2018.

\bibitem[Madry et~al.(2018)]{MadryMSTV18}
Aleksander Madry et~al.
\newblock Towards deep learning models resistant to adversarial attacks.
\newblock In \emph{ICLR 2018}, 2018.

\bibitem[Mahabadi et~al.(2020)]{MahabadiBH20}
Rabeeh~Karimi Mahabadi et~al.
\newblock End-to-end bias mitigation by modelling biases in corpora.
\newblock In \emph{ACL 2020}, pages 8706--8716. ACL, 2020.

\bibitem[Mansour et~al.(2009)]{MansourMR09}
Yishay Mansour et~al.
\newblock Domain adaptation: Learning bounds and algorithms.
\newblock In \emph{{COLT} 2009 - The 22nd Conference on Learning Theory,
  Montreal, Quebec, Canada, June 18-21, 2009}, 2009.

\bibitem[Marblestone et~al.(2016)]{MarblestoneWK16}
Adam~H. Marblestone et~al.
\newblock Toward an integration of deep learning and neuroscience.
\newblock \emph{Frontiers Comput. Neurosci.}, 10:\penalty0 94, 2016.

\bibitem[McDonald(2014)]{mcdonald2014confounding}
JH~McDonald.
\newblock Confounding variables.
\newblock \emph{Handbook of biological statistics, 3rd edn. Sparky House
  Publishing, Baltimore}, pages 24--28, 2014.

\bibitem[Motiian et~al.(2017)]{motiian2017unified}
Saeid Motiian et~al.
\newblock Unified deep supervised domain adaptation and generalization.
\newblock In \emph{ICCV}, volume~2, page~3, 2017.

\bibitem[Nam et~al.(2020)]{nam2020learning}
Junhyun Nam et~al.
\newblock Learning from failure: Training debiased classifier from biased
  classifier.
\newblock In \emph{NeurIPS}, 2020.

\bibitem[Namkoong and Duchi(2016)]{NIPS2016_4588e674}
Hongseok Namkoong and John~C Duchi.
\newblock Stochastic gradient methods for distributionally robust optimization
  with f-divergences.
\newblock In \emph{NeurIPS}, 2016.

\bibitem[Nayebi and Ganguli(2017)]{nayebi2017biologically}
Aran Nayebi and Surya Ganguli.
\newblock Biologically inspired protection of deep networks from adversarial
  attacks.
\newblock \emph{arXiv:1703.09202}, 2017.

\bibitem[Ribeiro et~al.(2016)]{ribeiro2016should}
Marco~Tulio Ribeiro et~al.
\newblock " why should i trust you?" explaining the predictions of any
  classifier.
\newblock In \emph{KDD 2017=6}, pages 1135--1144, 2016.

\bibitem[Ribeiro et~al.(2018)]{ribeiro2018anchors}
Marco~Tulio Ribeiro et~al.
\newblock Anchors: High-precision model-agnostic explanations.
\newblock In \emph{AAAI}, volume~32, 2018.

\bibitem[Ross et~al.(2017)]{ross2017right}
Andrew~Slavin Ross et~al.
\newblock Right for the right reasons: Training differentiable models by
  constraining their explanations.
\newblock \emph{arXiv:1703.03717}, 2017.

\bibitem[Rozantsev et~al.(2018)]{rozantsev2018beyond}
Artem Rozantsev et~al.
\newblock Beyond sharing weights for deep domain adaptation.
\newblock \emph{IEEE transactions on pattern analysis and machine
  intelligence}, 41\penalty0 (4):\penalty0 801--814, 2018.

\bibitem[Sagawa* et~al.(2020)Sagawa*, Koh*,
  et~al.]{Sagawa*2020Distributionally}
Shiori Sagawa*, Pang~Wei Koh*, et~al.
\newblock Distributionally robust neural networks.
\newblock In \emph{ICLR}, 2020.

\bibitem[Schramowski et~al.(2020)]{schramowski2020making}
Patrick Schramowski et~al.
\newblock Making deep neural networks right for the right scientific reasons by
  interacting with their explanations.
\newblock \emph{Nature Machine Intelligence}, 2\penalty0 (8):\penalty0
  476--486, 2020.

\bibitem[Shankar et~al.(2018)]{ShankarPCCJS18}
Shiv Shankar et~al.
\newblock Generalizing across domains via cross-gradient training.
\newblock In \emph{6th ICLR, {ICLR} 2018}, 2018.

\bibitem[Sun et~al.(2019)]{SunGTHEZMBCW19}
Tony Sun et~al.
\newblock Mitigating gender bias in natural language processing: Literature
  review.
\newblock In \emph{ACL 2019}, pages 1630--1640. ACL, 2019.

\bibitem[Szegedy et~al.(2013)]{szegedy2013intriguing}
Christian Szegedy et~al.
\newblock Intriguing properties of neural networks.
\newblock \emph{arXiv:1312.6199}, 2013.

\bibitem[Teso and Kersting(2019)]{teso2019explanatory}
Stefano Teso and Kristian Kersting.
\newblock Explanatory interactive machine learning.
\newblock In \emph{Proceedings of the 2019 AAAI/ACM Conference on AI, Ethics,
  and Society}, pages 239--245, 2019.

\bibitem[Torralba and Efros(2011)]{torralba2011unbiased}
Antonio Torralba and Alexei~A Efros.
\newblock Unbiased look at dataset bias.
\newblock In \emph{CVPR 2011}, pages 1521--1528. IEEE, 2011.

\bibitem[Vigen(2015)]{vigen2015spurious}
Tyler Vigen.
\newblock \emph{Spurious correlations}.
\newblock Hachette books, 2015.

\bibitem[Wang et~al.(2019{\natexlab{a}})]{WangGLX19}
Haohan Wang et~al.
\newblock Learning robust global representations by penalizing local predictive
  power.
\newblock In \emph{NeurIPS 2019}, pages 10506--10518, 2019{\natexlab{a}}.

\bibitem[Wang et~al.(2019{\natexlab{b}})]{WangHLX19}
Haohan Wang et~al.
\newblock Learning robust representations by projecting superficial statistics
  out.
\newblock In \emph{ICLR 2019}. OpenReview.net, 2019{\natexlab{b}}.

\bibitem[Wang et~al.(2020)]{wang2020high}
Haohan Wang et~al.
\newblock High-frequency component helps explain the generalization of
  convolutional neural networks.
\newblock In \emph{CVPR}, pages 8684--8694, 2020.

\bibitem[Wang et~al.(2022)]{wang2020squared}
Haohan Wang et~al.
\newblock Toward learning robust and invariant representations with alignment
  regularization and data augmentation.
\newblock \emph{KDD}, 2022.

\bibitem[Yoon et~al.(2019)]{yoon2018invase}
Jinsung Yoon et~al.
\newblock {INVASE}: Instance-wise variable selection using neural networks.
\newblock In \emph{ICLR}, 2019.

\bibitem[Zhang et~al.(2017)]{zhang2017mixup}
Hongyi Zhang et~al.
\newblock mixup: Beyond empirical risk minimization.
\newblock \emph{arXiv:1710.09412}, 2017.

\bibitem[Zhang et~al.(2019)]{ZhangLLJ19}
Yuchen Zhang et~al.
\newblock Bridging theory and algorithm for domain adaptation.
\newblock In \emph{ICML, 2019}, volume~97, pages 7404--7413. {PMLR}, 2019.

\end{thebibliography}

\appendix
\newpage 
\onecolumn

\newpage 
\section{Proofs of Theoretical Discussions}

\subsection{Lemma A.1 and Proof}

\begin{lemma}
With sample $(\x, \y)$ and two labeling functions $f_1(\x)=f_2(\x)=\y$, 
for an estimated $\theta \in \Theta$, if $\theta(\x)=\y$, then with 
\textbf{A3} , we have
\begin{align}
    d(\theta, f_1, \x) = 1 \implies r(\theta, \mathcal{A}(f_2,\x))=1. 
\end{align}
\label{lemma:iff}
\end{lemma}
\begin{proof}
If $\theta(\x)=\y$
and $d(\theta, f_1, \x) = 1$, according to \textbf{A3}, we have
$d(\theta, f_2, \x) = 0$. 

We prove this by contradiction. 

If the conclusion does not hold, 
$r(\theta, \mathcal{A}(f_2,\x)) = 0$, 
which means 
\begin{align}
   \max_{\x_{\mathcal{A}(f_2,\x)} \in \mathcal{X}_{\mathcal{A}(f_2,\x)}} |\theta(\x) - \y| = 0
\end{align}
Together with $d(\theta, f_2, \x) = 0$, which means
\begin{align}
    \max_{\z \in \mathcal{X}: \z_{\mathcal{A}(f_2,\x)}=\x_{\mathcal{A}(f_2,\x)}} |\theta(\z) - \y| = 0, 
\end{align} 
we will have
\begin{align}
    \max_{\x \in \mathcal{X}} |\theta(\x) - \y| = 0, 
\end{align}
which is $\theta(\x) = \y$ for any $\x \in \mP$. 

This contradicts with the premises in \textbf{A3} ($\theta$ is not a constant function). 


\end{proof}

\subsection{Theorem 3.1 and Proof}
\begin{theorem*}
With Assumptions \textbf{A1}-\textbf{A3}, with probability at least $1 - \delta$, we have 
\begin{align}
    \eptt \leq \wepst + c(\theta) + \phi(|\Theta|, n,  \delta)
\end{align}
where $c(\theta) =  \dfrac{1}{n}\sum_{(\x, \y) \in (\X, \Y)_{\mP_s}} \mathbb{I}[\theta(\x)=\y]r(\theta, \mathcal{A}(\s,\x))$. 
\end{theorem*}

\begin{proof}

\begin{align}
    \wepst =& \dfrac{1}{n} \sum_{(\x, \y) \in (\X, \Y)_{\mP_s}} |\theta(\x)-f(\x)| \\
    =& 1 - \dfrac{1}{n} \sum_{(\x, \y) \in (\X, \Y)_{\mP_s}} \big( 
    \mathbb{I}[\theta(\x)=f(\x)]
    \big)\\
    =& 1 - \dfrac{1}{n} \sum_{(\x, \y) \in (\X, \Y)_{\mP_s}} \big( 
    \mathbb{I}[\theta(\x)=f(\x)]\mathbb{I}[d(\theta, \h,\x)=0] + 
    \mathbb{I}[\theta(\x)=f(\x)]\mathbb{I}[d(\theta, \h,\x)=1]
    \big) \\
    =& 1 - \dfrac{1}{n} \sum_{(\x, \y) \in (\X, \Y)_{\mP_s}} \big(
    \mathbb{I}[\theta(\x)=f(\x)]\mathbb{I}[d(\theta, \h,\x)=0]
    \big) -
    \dfrac{1}{n} \sum_{(\x, \y) \in (\X, \Y)_{\mP_s}} \mathbb{I}[\theta(\x)=f(\x)]\mathbb{I}[d(\theta, \h,\x)=1] \\
    \geq & \widehat{\epsilon}_h(\theta) - 
    \dfrac{1}{n} \sum_{(\x, \y) \in (\X, \Y)_{\mP_s}} 
    \mathbb{I}[\theta(\x)=f(\x)]r(\theta, \mathcal{A}(\s,\x)), 
\end{align}
where the last line used Lemma~\ref{lemma:iff}. 

Thus, we have
\begin{align}
    \widehat{\epsilon}_h(\theta) \leq \widehat{\epsilon}(\theta) + 
    \dfrac{1}{n} \sum_{(\x, \y) \in (\X, \Y)_{\mP_s}}
    \mathbb{I}[\theta(\x)=f(\x)]r(\theta, \mathcal{A}(\s,\x)) 
\end{align}
where 
\begin{align}
    \widehat{\epsilon}_h(\theta) = 1 - \dfrac{1}{n} \sum_{(\x, \y) \in (\X, \Y)_{\mP_s}} \big(
    \mathbb{I}[\theta(\x)=f(\x)]\mathbb{I}[d(\theta, \h,\x)=0]
    \big), 
\end{align}
which describes the correctly predicted terms that $\theta$ functions the same as $\h$ and all the wrongly predicted terms. Therefore, conventional generalization analysis through uniform convergence applies, and we have
\begin{align}
    \eptt \leq \widehat{\epsilon}_h(\theta)  + \phi(|\Theta|, n,  \delta)
\end{align}

Thus, we have:
\begin{align}
    \eptt \leq  \wepst + 
    \dfrac{1}{n} \sum_{(\x, \y) \in (\X, \Y)_{\mP_s}} 
    \mathbb{I}[\theta(\x)=\y]r(\theta, \mathcal{A}(\s,\x)) +
    \phi(|\Theta|, n,  \delta)
\end{align}
\end{proof}

\subsection{Theorem 3.2 and Proof}
\begin{theorem*}
With Assumptions \textbf{A2}-\textbf{A4}, 
and if $1 - \h \in \Theta$, 
we have
\begin{align}
   c(\theta) \leq D_\Theta(\mP_s, \mP_t) + \dfrac{1}{n}\sum_{(\x,\y) \in (\X, \Y)_{\mP_t}} \mathbb{I}[\theta(\x)=\y]r(\theta, \mathcal{A}(\s,\x))
\end{align}
where 
$c(\theta) =  \dfrac{1}{n}\sum_{(\x,\y) \in (\X, \Y)_{\mP_s}} \mathbb{I}[\theta(\x)=\y]r(\theta, \mathcal{A}(\s,\x))$
and $D_\Theta(\mP_s, \mP_t)$ is defined as in~\eqref{eq:h-divergence}. 
\end{theorem*}

\begin{proof}
By definition, $g(\x) \in \Theta\Delta\Theta \iff g(\x)=\theta(\x)\oplus\theta'(\x)$ for some $\theta, \theta' \in \Theta$, 
together with Lemma 2 and Lemma 3 of \citep{ben2010theory}, 
we have
\begin{align}
    D_\Theta(\mP_s, \mP_t) =& \dfrac{1}{n}\max_{\theta, \theta' \in \Theta}
    \big\vert 
    \sum_{(\x,\y) \in (\X,\Y)_{\mP_s}}\vert\theta(\x)-\theta'(\x)\vert
    -
    \sum_{(\x,\y) \in (\X,\Y)_{\mP_t}}\vert\theta(\x)-\theta'(\x)\vert
    \big\vert \\
    \geq & \dfrac{1}{n} \big\vert 
    \sum_{(\x,\y) \in (\X,\Y)_{\mP_s}}\vert\theta(\x)-f_z(\x)\vert 
    -
    \sum_{(\x,\y) \in (\X,\Y)_{\mP_t}}\vert\theta(\x)-f_z(\x)\vert 
    \big\vert  \\
    = & \dfrac{1}{n} \big\vert 
    \sum_{(\x,\y) \in (\X,\Y)_{\mP_s}}\mathbb{I}[\theta(\x)=\y] 
    -
    \sum_{(\x,\y) \in (\X,\Y)_{\mP_t}}\mathbb{I}[\theta(\x)=\y] 
    \big\vert \\
    = & \dfrac{1}{n} \big\vert 
    \sum_{(\x,\y) \in (\X,\Y)_{\mP_s}}\mathbb{I}[\theta(\x)=\y]\mathbb{I}[r(\theta, \mathcal{A}(\s,\x))=1]
    -
    \sum_{(\x,\y) \in (\X,\Y)_{\mP_t}}\mathbb{I}[\theta(\x)=\y]\mathbb{I}[r(\theta, \mathcal{A}(\s,\x))=1] \\
    & + \sum_{(\x,\y) \in (\X,\Y)_{\mP_s}}\mathbb{I}[\theta(\x)=\y]\mathbb{I}[r(\theta, \mathcal{A}(\s,\x))=0]
    -
    \sum_{(\x,\y) \in (\X,\Y)_{\mP_t}}\mathbb{I}[\theta(\x)=\y]\mathbb{I}[r(\theta, \mathcal{A}(\s,\x))=0]
    \big\vert \\
    = & \dfrac{1}{n} \big\vert 
    \sum_{(\x,\y) \in (\X,\Y)_{\mP_s}}\mathbb{I}[\theta(\x)=\y]r(\theta, \mathcal{A}(\s,\x)) 
    -
    \sum_{(\x,\y) \in (\X,\Y)_{\mP_t}}\mathbb{I}[\theta(\x)=\y]r(\theta, \mathcal{A}(\s,\x))
    \big\vert \\
    \geq & c(\theta) - 
    \sum_{(\x,\y) \in (\X,\Y)_{\mP_t}}\mathbb{I}[\theta(\x)=\y]r(\theta, \mathcal{A}(\s,\x))
\end{align}

First line: see Lemma 2 and Lemma 3 of \citep{ben2010theory}. 

Second line: if $1-\h \in \Theta$, and we use $f_z$ to denote $1-\h$. 

Fifth line is a result of using that fact that 
\begin{align}
    \sum_{(\x,\y) \in (\X,\Y)_{\mP_s}}\mathbb{I}[\theta(\x)=\y]\mathbb{I}[r(\theta, \mathcal{A}(\s,\x))=0]
    =
    \sum_{(\x,\y) \in (\X,\Y)_{\mP_t}}\mathbb{I}[\theta(\x)=\y]\mathbb{I}[r(\theta, \mathcal{A}(\s,\x))=0]
    \label{eq:thm:da:proof:5}
\end{align}
as a result of our assumptions. 
Now we present the details of this argument:

According to \textbf{A3}, if $\theta(\x)=\y$, $d(\theta,\h,\x)d(\theta,\s,\x)=0$. 
Since $r(\theta, \mathcal{A}(\s,\x))=0$, $d(\theta,\s,\x)$ cannot be 0 unless $\theta$ is a constant mapping that maps every sample to 0 (which will contradicts \textbf{A3}). 
Thus, we have $d(\theta,\h,\x)=0$. 

Therefore, we can rewrite the left-hand term following
\begin{align}
    \sum_{(\x,\y) \in (\X,\Y)_{\mP_s}}\mathbb{I}[\theta(\x)=\y]\mathbb{I}[r(\theta, \mathcal{A}(\s,\x))=0]
    = 
    \sum_{(\x,\y) \in (\X,\Y)_{\mP_s}}\mathbb{I}[\theta(\x)=\y]\mathbb{I}[d(\theta,\h,\x)=0]
\end{align}
and similarly
\begin{align}
    \sum_{(\x,\y) \in (\X,\Y)_{\mP_t}}\mathbb{I}[\theta(\x)=\y]\mathbb{I}[r(\theta, \mathcal{A}(\s,\x))=0]
    = 
    \sum_{(\x,\y) \in (\X,\Y)_{\mP_t}}\mathbb{I}[\theta(\x)=\y]\mathbb{I}[d(\theta,\h,\x)=0]
\end{align}

We recap the definition of $d(\cdot, \cdot, \x)$, thus $d(\theta,\h,\x)=0$ means
\begin{align}
    d(\theta, \h,\x) = \max_{\z \in \mathcal{X}: \z_{\mathcal{A}(f,\x)}=\x_{\mathcal{A}(\h,\x)}} |\theta(\z) - \h(\z)| = 0
\end{align}
Therefore $d(\theta,\h,\x)=0$ implies $\mathbb{I}(\theta(\x)=\y)$, 
and 
\begin{align}
    |\theta(\z) - \h(\z)| = 0 \quad \forall \quad  \z_{\mathcal{A}(\h,\x)} = \x_{\mathcal{A}(\h,\x)}
\end{align}

Therefore, we can continue to rewrite the left-hand term following
\begin{align}
    \sum_{(\x,\y) \in (\X,\Y)_{\mP_s}}\mathbb{I}[\theta(\x)=\y]\mathbb{I}[d(\theta,\h,\x)=0]
    =
    \sum_{(\x,\y) \in (\X,\Y)_{\mP_s}}
    \mathbb{I}[\theta(\z) - \h(\z)]
    =\sum_{(\x,\y) \in (\X,\Y)_{\mP_s}}
    \mathbb{I}[\theta(\x) - \h(\x)]
\end{align}
and similarly
\begin{align}
    \sum_{(\x,\y) \in (\X,\Y)_{\mP_t}}\mathbb{I}[\theta(\x)=\y]\mathbb{I}[d(\theta,\h,\x)=0]
    =
    \sum_{(\x,\y) \in (\X,\Y)_{\mP_t}}
    \mathbb{I}[\theta(\z) - \h(\z)]
\end{align}
where $\z$ denotes any $\z \in \mathcal{X}$ and $\z_{\mathcal{A}(\h,\x)} = \x_{\mathcal{A}(\h,\x)}$. 

Further, because of \textbf{A4}, we have
\begin{align}
    \sum_{(\x,\y) \in (\X,\Y)_{\mP_t}}
    \mathbb{I}[\theta(\z) - \h(\z)]
    =\sum_{(\x,\y) \in (\X,\Y)_{\mP_s}}
    \mathbb{I}[\theta(\x) - \h(\x)].
\end{align}
Thus, we show the \eqref{eq:thm:da:proof:5} holds 
and conclude our proof. 
\end{proof}

\newpage
\section{Additional Discussion to Connect to Robust Machine Learning Methods}

\subsection{Worst-case Data Augmentation in Practice}
In practice, when we use data augmentation to learn human-aligned models, we need either of the two following assumptions to hold: 
\begin{itemize}
    \item [\textbf{A4-1}:] \textbf{Labeling Functions Separability of Features} 
    For any $\x \in \mathcal{X}$, $\mathcal{A}(\h,\x) \cap \mathcal{A}(\s,\x) = \emptyset$
    \item [\textbf{A4-2}:] \textbf{Labeling Functions Separability of Input Space}
    We redefine $\s: \dom(\s)\rightarrow \mathcal{Y}$ and $\dom(\s) \subsetneq \mathcal{X}$.
    For any $\x \in \mathcal{X}$,
    $\max_{\z \in \dom(\s)\cap\dom(\h)}|\h(\z) - \h(\z)| = 0$
\end{itemize}

While both of these assumptions appear strong, 
we believe a general discussion of human-aligned models may not be 
able to built without these assumptions. 
In particular, \textbf{A4-1} describes the situations that 
$\h'$ do not use the same set of features as $\s'$. 
One example of this situation could be that 
the background of an image in dog vs. cat classification is considered features for $\s'$, 
and the foreground of an image is considered as features for $\h'$. 
\textbf{A4-2} describes the situations that 
while $\s'$ can uses the features that are considered by $\h'$, 
the perturbation of the features within the domain of $\s'$
will not change the output of $\h'$. 
One example of this situation could be that 
the texture of dog or cat in the dog vs. cat classification, 
while the texture can be perturbed,  
the perturbation cannot be allowed to an arbitrary scale of pixels 
(otherwise the perturbation is not a perturbation of texture). 
If neither of these assumptions holds, 
then the perturbation will be allowed to replace a dog's body with the one of a dolphin, 
and even human may not be able to confidently decide the resulting image is a dog, 
thus human-aligned learning will not be worth discussion. 

\subsection{Derivation of Weighted Risk Minimization.}



\subsubsection{Connections to Distributionally Robust Optimization (DRO)}
Recall that we generalize the above analysis of worst-case data augmentation to a DRO problem \citep{ben2013robust, duchi2021statistics}. Given $n$ data points, consider a perturbation set $\mathcal{Q} := \{ \mathbf{x}_{\mathcal{A}(f_m, \mathbf{x}_i)} \in \operatorname{dom}(f)_{\mathcal{A}(f_m, \mathbf{x}_i)} \}_{i=1}^n$ encoding the features of $\x$ indexed by $\mathcal{A}(f,\x)$ over input space $\dom(f_m)$. Denote $q(\x, \y)$ and $p(\x,\y)$ are densities from the $\mathcal{Q}$ and training distribution $\mathcal{X} \times \mathcal{Y}$, respectively. Then (\ref{eq:method:worst-da-result}) can be rewritten as a DRO problem over a new distribution $\mathcal{Q}$.

\begin{equation}
    c(\theta) \le \min_{\theta \in \Theta} \max_{\z\in \mathcal{Q}(\x)} \dfrac{1}{n}\sum_{(\x,\y)\in(\mathbf{X},\mathbf{Y})} \ell(\theta(\z),\y)
    \label{eq:method:dro}
\end{equation}

To transform DRO into WRM, we introduce the following assumptions about perturbation set $\mathcal{Q}$:
\begin{itemize}
    \item [\textbf{A2-1}:] \textbf{$q \ll p$.} $p(x, y) = 0 \implies q(x, y) = 0$
    \item [\textbf{A2-2}:] \textbf{$f$-Divergence.} Given a function $\xi$ is convex and $\xi(1) = 0$ and $\delta > 0$ as a radius to control the degree of the distribution shift, $D_{\xi}\left(q\left(\mathbf{x}, \mathbf{y}\right) \| p\left(\mathbf{x}, \mathbf{y}\right)\right) \leq \delta$ holds.
\end{itemize}

$\mathcal{Q}$ encodes the priors about feature perturbation that model should be robust to. Therefore, choosing $f$-divergence as the distance metric where $\xi$ is convex with $\xi(1) = 0$, $\delta > 0$ as a radius to control the degree of the distribution shift, adversarial robustness in Section~\ref{sec:worst-da} can be viewed as an example of DRO on an infinite family of distributions with implicit assumptions that samples in $\mathcal{Q}$ are visually indistinguishable from original ones. For $p$ and $q$ that $p(\x, \y) = 0$ implies $q(\x, \y) = 0$, we arrive at a generic weighted risk minimization (WRM) formulation \citep{NIPS2016_4588e674, duchi2021statistics} when weights (by default as density ratios) $\lambda = q(\x, \y)/p(\x,\y)$ in (\ref{eq:method:wrm_loss}) derived from misaligned functions for 
\begin{align}
\begin{split}
    c(\theta) \le \min _{\theta \in \Theta} \max _{\z \in \mathcal{Q}_\xi(\x)} \frac{1}{n} \sum_{(\x,\y)\in(\mathbf{X},\mathbf{Y})} \lambda\left(\z\right) \cdot \ell\left(\theta\left(\z\right), \y\right)
    \label{eq:method:wrm_loss}
\end{split}
\end{align}


where the uncertainty set $\mathcal{Q}_\xi$ is reformulated as 
\begin{align}
\mathcal{Q}_\xi := \{\lambda(\z_i)| & D_\xi(q||p) \le \delta, \\ & \sum_{i=1}^{n}\lambda(\z_i)=1, \\ & \forall \lambda(\z_i)\ge 0 \}
\end{align}

When $\lambda(\cdot)=q(\mathbf{x}, \mathbf{y}) / p(\mathbf{x}, \mathbf{y})$ is the density ratio, we use change of measure technique to show the equivalence of DRO and WRM by transoforming the optimization problem on $q$ to an optimization problem $\lambda(\cdot)$. 
And the inner optimization problem are equivalent to
\begin{equation}
    \mathbb{E}_q[\ell(\theta,\x)] = \int \ell(\theta,\x) q(\z)d\z = \int \ell(\theta,\x) \frac{q(\z)}{p(\z)}p(\z)d\z=\mathbb{E}_{p}[\lambda(\x)\ell(\theta,\x)]
\end{equation}

Moreover, choosing $f=x\log(x)$, $f$-divergence becomes KL-divergence and then the constraint can be converted to 
\begin{equation}
    D_\xi(p \| q)=\int_{q>0} \frac{p(\x)}{q(\x)} \log\left(\frac{p(\x)}{q(\x)} \right) q(\x) d\x
    = \mathbb{E}_q[\lambda(\x)\log\lambda(\x)] \le \delta
\end{equation}

Next we prove the equivalence of DRO and WRM for general $\mathbf{\lambda}$ under additional assumptions below. 
\begin{itemize}
    \item [\textbf{A2-3}:] \textbf{Finite perturbation set.} $\mathcal{Q}$ is a finite set.
    \item [\textbf{A2-4}:] \textbf{Convexity.} Loss function $\ell$ is convex in $\theta$ and concave in $\lambda$. $\mathcal{Q}$ and $\Theta$ are convex sets.
    \item [\textbf{A2-5}:] \textbf{Continuity.} Loss function $\ell$ and its weighted sum $\sum\limits_{(\x,\y)\in(\mathbf{X}, \mathbf{Y})} \lambda(\mathbf{z}) \ell\left(\theta\left(\mathbf{z}\right), \mathbf{y}\right)$ are continuous.
    \item [\textbf{A2-6}:] \textbf{Compactness.} $\mathcal{Q}$ and $\Theta$ are compact.
\end{itemize}

Given $n$ data points, we introduce slack variable $\xi$ and consider a constrained optimization formulation of (\ref{eq:method:dro}) as
\begin{equation}
    \min_{\theta \in \Theta, \xi} \xi \quad \quad s.t. \quad \sum_{i=1}^{n}  \lambda(\z_i)\ell(\theta(\z_i), \y_i)-\xi \le 0 \quad \forall \z \in \mathcal{Q}
\end{equation}

By the strong convex duality, we have the Lagrangian
$L(\theta,\alpha, \lambda(\z_1), \dots, \lambda(\z_n)) = \alpha+\sum\limits_{i=1}^n \lambda(\z_i)(\ell(\theta(\z_i),\y_i)-\alpha)$ and the dual problem as

\begin{equation}
    \max_{\forall\lambda_i\ge0} \min_{\theta,\alpha}L(\theta, \alpha, \lambda(\z_1), \dots, \lambda(\z_n)) \quad s.t. \sum\limits_{i=1}^{n}\lambda(\z_i)=1,\quad i=1,\dots,n
\end{equation}
Which can be expressed as

\begin{equation}
    \max_{\forall\mathbf{\lambda}\succeq0} \min_{\theta\in \Theta} \mathcal{L}(\mathbf{\lambda},\theta) = \max_{\forall\lambda_i\succeq0} \min_{\theta\in \Theta} \sum_{i=1}^{n} \lambda_i \ell(\theta(\z_i), \y_i) \quad s.t. \sum\limits_{i=1}^{n}\lambda(\z_i)=1, i=1,\dots,n
\end{equation}

By the minimax equality, we have
\begin{equation}
    \max_{\forall\mathbf{\lambda}\succeq0} \min_{\theta\in \Theta} \mathcal{L}(\mathbf{\lambda},\theta) = \min_{\theta\in \Theta} \max_{\forall\mathbf{\lambda}\succeq0}  \mathcal{L}(\mathbf{\lambda},\theta)
\end{equation}
Denote the optimality of $\max\limits_{\forall\mathbf{\lambda}\succeq0} \min\limits_{\theta\in \Theta} \mathcal{L}(\mathbf{\lambda},\theta)$ and $\min\limits_{\theta\in \Theta} \max\limits_{\forall\mathbf{\lambda}\succeq0}  \mathcal{L}(\mathbf{\lambda},\theta)$ as $\mathbf{\lambda}^*$ and $\theta^*\in \Theta$, respectively. Then we have $(\mathbf{\lambda}^*, \theta^*)$ form a saddle point that
\begin{equation}
    \max_{\mathbf{\lambda}\succeq0} \mathcal{L}(\theta^*,\mathbf{\lambda})
    = \mathcal{L}(\theta^*,\mathbf{\lambda}^*) = \min_{\theta \in \Theta} \mathcal{L}(\theta,\mathbf{\lambda}^*)
\end{equation}
which means that $\mathbf{\lambda}^*$ exists in the WRM such that $\theta^* \in \argmin_\theta \mathcal{L}(\mathbf{\lambda}, \theta)$ is optimal for DRO.

Intuitively, learner $\theta$ and adversary $\phi$ are playing a minimax game where $\phi$ finds worst-case weights and computationally-identifiable regions of errors to improve the robustness of the learner $\theta$. In this scenario, we unify a line of WRM approaches where weights $\lambda$ are mainly determined by misaligned features $\mathcal{A}(f_m, \mathbf{x})$, either parameterized by a biased model or derived from some heuristic statistics. 

\subsection{Details to Connect Methods to Regularize the Hypothesis Space}

First, we need to formally introduce the properties regarding $\s'$, as a correspondence to those of $\s$. 

\paragraph{Notations and Background with Encoder/Decoder Structure}
With the same binary classification problem 
from feature space $\mathcal{X}$ to 
label space $\mathcal{Y}$. 
We consider the encoder $\theta_e: \mathcal{X} \rightarrow \mathcal{E}$
and decoder $\theta_d: \mathcal{E} \rightarrow \mathcal{Y}$, 
$f': \mathcal{E} \rightarrow \mathcal{Y}$ is the function that maps the embedding to the label. 

Similarly, we introduce the assumptions on the $\mathcal{E}$ space. 
\begin{itemize}
    \item [\textbf{A2'}:] \textbf{Existence of Superficial Features:}
    For any $\x \in \mathcal{X}$ and an oracle encoder $\theta_e$ that $\e=\theta_e(\x)$, $\y := \h'(\e)$. 
    We also have a $\s'$ 
    that is different from $\h'$, and for $\x \sim \mP_s$ and $\e = \theta_e(\x)$, 
    $\h'(\e) = \s'(\e)$. 
\end{itemize}



\begin{itemize}
    \item [\textbf{A3'}:] \textbf{Realized Hypothesis:} 
    Given a large enough hypothesis space $\Theta_d$ for decoders, for any sample $(\x, \y)$ and an encoder $\theta_e$ that $\e=\theta_e(\x)$, 
    for any $\theta_d \in \Theta_d$, 
    which is not a constant mapping, 
    if $\theta_d(\e)=\y$, then 
    $d(\theta_d, \h', \e)d(\theta_d, \s', \e)=0$
\end{itemize}

With the above assumptions, following the same logic, 
we can derive the theorem corresponding to Theorem 3.1, 
with the only difference that how $c(\theta)$ is now derived. 

\begin{lemma}
With Assumptions \textbf{A1}, \textbf{A2'}, \textbf{A3'}, $l(\cdot, \cdot)$ is a zero-one loss, with probability as least $1 - \delta$, we have 
\begin{align}
    \eptt \leq \wepst + c(\theta) + \phi(|\Theta|, n,  \delta)
\end{align}
where $c(\theta) =  \dfrac{1}{n}\sum_{(\x, \y) \in (\X, \Y)_{\mP_s}} \mathbb{I}[\theta(\x)=\y]r(\theta_d, \mathcal{A}(\s',\theta_e(\x)))$. 
\end{lemma}

Now, we continue to show that how training for small $c(\theta)$ amounts to solving \eqref{eq:regularization}. 
To proceed, we need either of the two following assumptions to hold: 
\begin{itemize}
    \item [\textbf{A4-1'}:] \textbf{Labeling Functions Separability of Features} 
    For any $\x \in \mathcal{X}$ and an encoder $\theta_e$ that $\e=\theta_e(\x)$, $\mathcal{A}(\h',\e) \cap \mathcal{A}(\s',\e) = \emptyset$
    \item [\textbf{A4-2'}:] \textbf{Labeling Functions Separability of Input Space}
    We redefine $\s': \dom(\s') \rightarrow \mathcal{Y}$ and $\dom(\s') \subsetneq \mathcal{E}$.  
    For any $\x \in \mathcal{X}$ and an encoder $\theta_e$ that $\e=\theta_e(\x)$,
    $\max_{\z \in \dom(\s')\cap\dom(\h')}|\h'(\z) - \h'(\z)| = 0$
\end{itemize}

Also, notice that, assumptions \textbf{A4-1'} and \textbf{A4-2'} also regulates
the encoder to be reasonably good. 
In other words, these assumptions will not hold for arbitrary encoders. 

Now, we continue to derive \eqref{eq:regularization} from Lemma B.1 as the following:
\begin{align*}
    c(\theta) 
    &= \dfrac{1}{n}\sum_{(\x,\y)\in (\X,\Y)} \mathbb{I}[\theta_d(\theta_e(\x))=\y] r(\theta_d, \mathcal{A}(\s',\x))  \\
    &= \dfrac{1}{n}\sum_{(\x,\y)\in (\X,\Y)}
    \mathbb{I}[\theta_d(\theta_e(\x))=\y]
    \max_{\theta_e(\x)_{\mathcal{A}(\s',\x)} \in \dom(\theta_d)_{\mathcal{A}(\s',\x)}} |\theta_d(\theta_e(\x)) - \y| \\
    & = \dfrac{1}{n}\sum_{(\x,\y)\in (\X,\Y)}
    \max_{\theta_e(\x)_{\mathcal{A}(\s',\x)} \in \dom(\theta_d)_{\mathcal{A}(\s',\x)}} |\s'(\theta_e(\x)) - \y| \\ 
    & \leq \dfrac{1}{n}\sum_{(\x,\y)\in (\X,\Y)}\max_{\theta_e(\x) \in \dom(\theta_d)} |\s'(\theta_e(\x))- \y|
\end{align*}
The third line is because of the definition of $\mathbb{I}[\theta_d(\theta_e(\x))=\y] r(\theta_d, \mathcal{A}(\s',\x))$ and assumptions of \textnormal{A3'} and either \textbf{A4-1'} or \textbf{A4-2'}. 
Therefore, optimizing the empirical loss and $c(\theta)$ leads to 
\begin{align*}
    \min_{\theta_d, \theta_e}\dfrac{1}{n}\sum_{(\x,\y)\in (\X,\Y)}l(\theta_d(\theta_e(\x)), \y) - l(\s'(\theta_e(\x)), \y)
\end{align*}

\newpage 
\section{Theory-supporting Experiments}

\label{sec:app:exp:theory}

\paragraph{Synthetic Data with Spurious Correlation}
We extend the setup in Figure~\ref{fig:intro} to generate the synthetic dataset to test our methods. 
We study a binary classification problem over the data with $n$ samples and $p$ features, denoted as $\X \in \mathcal{R}^{n\times p}$. 
For every training and validation sample $i$, we generate feature $j$ as following:
\begin{align*}
    \X_j^{(i)} \sim \begin{cases}
    N(0, 1) & \textnormal{if}\; 1 \leq j \leq 3p/4 \\
    N(1, 1) & \textnormal{if}\; 3p/4 < j \leq p, \; \textnormal{and}\; y^{(i)}=1, \quad \textnormal{w.p.}\; \rho \\
    N(-1, 1) & \textnormal{if}\; 3p/4 < j \leq p, \; \textnormal{and}\; y^{(i)}=0, \quad \textnormal{w.p.}\; \rho \\
    N(0, 1) & \textnormal{if}\; 3p/4 < j \leq p, \quad \textnormal{w.p.}\; 1 - \rho
    \end{cases},
\end{align*}
In contrast, testing data are simply sampled with $\x_j^{(i)} \sim N(0, 1)$. 

To generate the label for training, validation, and test data, we sample two effect size vectors $\beta_1\in \mathcal{R}^{p/4}$ and $\beta_2\in \mathcal{R}^{p/4}$ whose each coefficient is sampled from a Normal distribution. We then generate two intermediate variables:
\begin{align*}
    \mathbf{c}_1^{(i)} = \X_{1, 2, \dots, p/4}^{(i)}\beta_1 \quad \textnormal{and} \quad \mathbf{c}_2^{(i)} = \X_{1, 2, \dots, p/4}^{(i)}\beta_2
\end{align*}
Then we transform these continuous intermediate variables into binary intermediate variables via Bernoulli sampling with the outcome of the inverse logit function ($g^{-1}(\cdot)$) over current responses, \textit{i.e.}, 
\begin{align*}
\mathbf{r}_1^{(i)} = \textnormal{Ber}(g^{-1}(\mathbf{c}_1^{(i)})) \quad \textnormal{and} \quad
\mathbf{r}_2^{(i)} = \textnormal{Ber}(g^{-1}(\mathbf{c}_2^{(i)}))
\end{align*}
Finally, the label for sample $i$ is determined as $\y^{(i)} = \mathbb{I}(\mathbf{r}_1^{(i)} = \mathbf{r}_2^{(i)})$, where $\mathbb{I}$ is the function that returns 1 if the condition holds and 0 otherwise. 

Intuitively, we create a dataset of $p$ features, half of the features are generalizable across train, validation and test datasets through a non-linear decision boundary, one-forth of the features are independent of the label, and the remaining features are spuriously correlated features: these features are correlated with the labels in train and validation set, but independent with the label in test dataset. There are about $\rho n$ train and validation samples have the correlated features. 

\begin{figure}[hbt]
    \centering
    \includegraphics[width=0.9\textwidth]{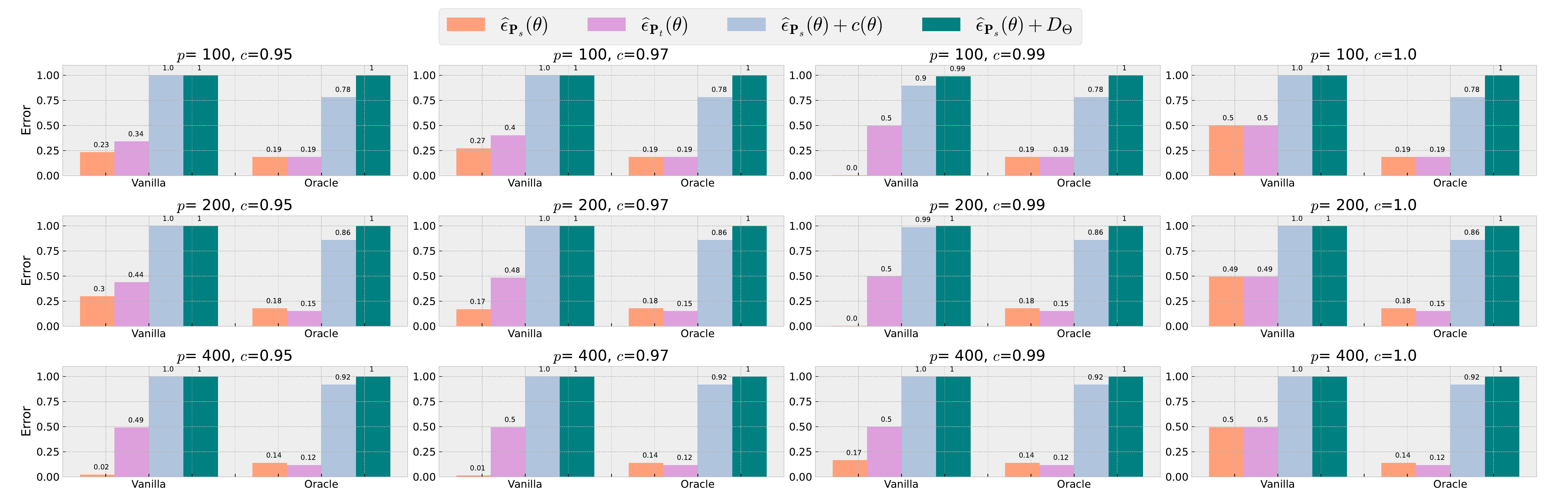}
    \caption{Results of Synthetic Data with Spurious Correlation. Each panel represents one setting. Five methods are reported in each panel. 
    For each method, four bars are plotted: from left to right, 
    $\wepst$, $\widehat{\epsilon}_{\mP_t}(\theta)$, $\wepst + c(\theta)$, and $\wepst + D_\Theta$.}
    \label{fig:synthetic}
\end{figure}

We train a vanilla ERM method, and in comparison, we also train an oracle method which that uses data augmentation to 
randomized the previously known spurious features.
We report training error (\textit{i.e.}, $\wepst$), test error (\textit{i.e.}, $\widehat{\epsilon}_{\mP_t}(\theta)$), 
$\wepst + c(\theta)$, and $\wepst + D_\Theta$
so that we can directly compare the bars to evaluate whether $c(\theta)$ can quantify the expected test error. 
Our results suggest that $c(\theta)$ is often a tighter estimation of the 
test error than $D_\Theta(\mP_s, \mP_t)$,
which aligns well with our analysis in Section~\ref{sec:cua}.


\begin{figure*}[h]
    \centering
    \small 
    \includegraphics[width=0.9\textwidth]{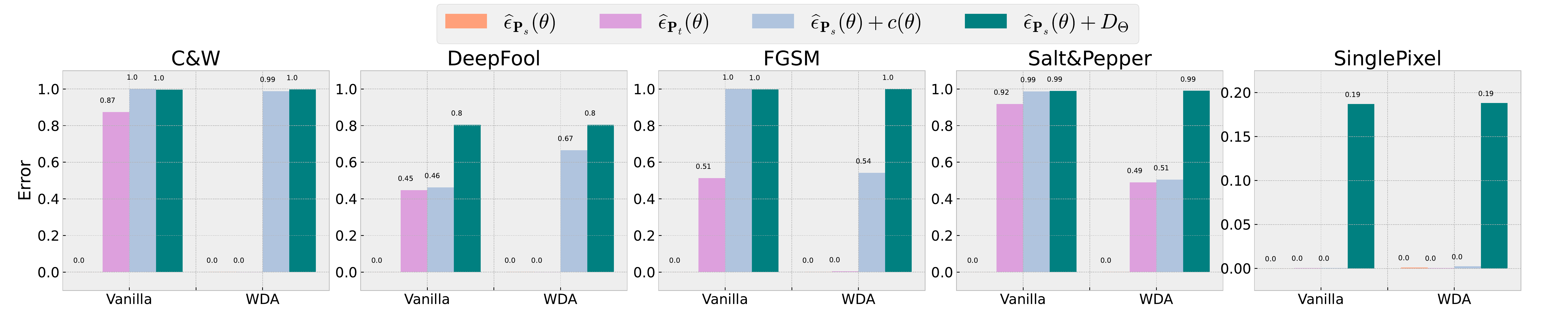}
    \caption{Binary MNIST classification error and estimated bounds. Each panel represents one out-of-domain data generated through an attack method. Four methods are reported in each panel. 
    For each method, four bars are plotted: from left to right, 
    $\wepst$, $\widehat{\epsilon}_{\mP_t}(\theta)$, $\wepst + c(\theta)$, and $\wepst + D_\Theta$. Some bars are not visible because the values are small. 
    }
    \label{fig:bMNIST}
\end{figure*}

\paragraph{Binary Digit Classification over Transferable Adversarial Examples}
For the second one, we consider a binary digit classification task, 
where the train and validation sets
are digits 0 and 1 from MNIST 
train and validation sets. 
To create the test set, 
we first estimate a model, 
and perform adversarial attacks over this model 
to generate the test samples with five adversarial attack 
methods (C\&W,
DeepFool,
FGSM, 
Salt\&Pepper, 
and SinglePixel). 
These adversarially generated examples 
are considered as the test set from another distribution. 

An advantage of this setup 
is that we can
have $\s$ well defined as
$1 - f_{adv}$, 
where the $f_{adv}$ is the function each adversarial attack relies on. 
Thus, according to our discussion on the estimation of $c(\theta)$ in Section~\ref{sec:cua}, 
we can directly use the corresponding adversarial attack methods to estimate $c(\theta)$ in our case.
Therefore, we can
assess our analysis on image classification.

We train the models with the vanilla method, and 
worst-case data augmentation (WDA, \textit{i.e.}, adversarial training). 
In addition to the training error (\textit{i.e.}, $\wepst$) and test error (\textit{i.e.}, $\widehat{\epsilon}_{\mP_t}(\theta)$), 
we also report $\wepst + c(\theta)$ and $\wepst + D_\Theta$
so that we can directly compare the bars to evaluate whether $c(\theta)$ can quantify the expected test error. 
By comparing the four different bars within every panel for every method, 
we notice that $c(\theta)$ is often a tighter estimation of the 
test error than $D_\Theta(\mP_s, \mP_t)$,
which aligns well with our analysis in Section~\ref{sec:cua}.

\end{document}